\RequirePackage{fix-cm}
\documentclass[smallextended]{svjour3}       
\smartqed  
\usepackage{natbib}
\usepackage[colorlinks=true,linkcolor=blue,citecolor=blue,urlcolor=blue]{hyperref}

\usepackage{graphicx}
\usepackage{multirow}%
\usepackage{amsmath,amssymb,amsfonts}%
\usepackage{mathrsfs}%
\usepackage[title]{appendix}%
\usepackage{xcolor}%
\usepackage{textcomp}%
\usepackage{manyfoot}%
\usepackage{booktabs}%
\usepackage{graphicx}
\usepackage{algorithm}%
\usepackage{algorithmicx}%
\usepackage{algpseudocode}%
\usepackage{listings}%
\usepackage{subcaption}
\usepackage{colortbl}
\usepackage{tcolorbox} 
\usepackage{enumitem}
\usepackage{marvosym}
\usepackage{tikz}
\usepackage{lineno} 

\definecolor{lightpurple}{RGB}{225, 213, 231}
\newcommand*\circled[1]{\tikz[baseline=(char.base)]{
            \node[shape=circle,draw,fill=lightpurple,inner sep=0.5pt, minimum size=1.2em] (char) {\fontsize{9}{12}\selectfont \textbf{#1}};}}

\definecolor{feature}{rgb}{0.6, 0.3, 0} 
\definecolor{lightgreen}{HTML}{ffffff}
\definecolor{lightgreenstrong}{HTML}{accca3}
\definecolor{green}{HTML}{71d156}
\newcommand{\bestmark}[1]{\includegraphics[height=5pt]{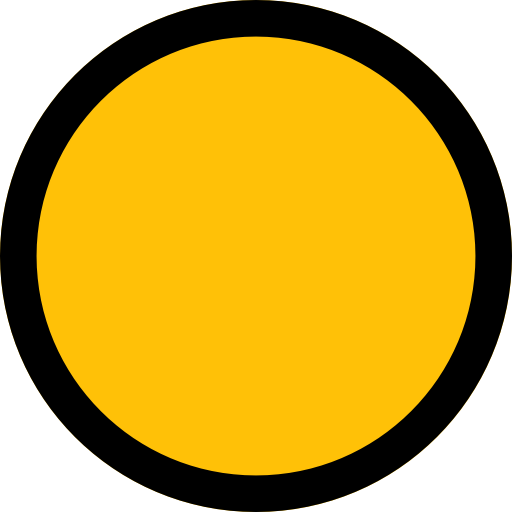} \textbf{#1}}

\newcommand*{\affaddr}[1]{#1} 
\newcommand*{\affmark}[1][*]{\textsuperscript{#1}}

\journalname{Empirical Software Engineering}
\begin{document}


\title{Leveraging Encoder-only Large Language Models for Mobile App Review Feature Extraction}

\titlerunning{Leveraging Encoder-only LLMs for Mobile App Review Feature Extraction}        

\author{Quim~Motger\affmark[1]         \and
        Alessio~Miaschi\affmark[2]     \and
        Felice~Dell'Orletta\affmark[2] \and
        Xavier~Franch\affmark[1]       \and
        Jordi~Marco\affmark[3]
}

\authorrunning{Quim Motger et al.} 

\institute{
            \Letter \quad  Quim Motger \\
              \email{joaquim.motger@upc.edu} \\\\
\affaddr{\affmark[1] Universitat Politècnica de Catalunya, Department of Service and Information System Engineering, Barcelona, Spain. E-mail: \{joaquim.motger,xavier.franch\}@upc.edu \\
\affmark[2] Institute for Computational Linguistics ``A. Zampolli" (ILC-CNR), ItaliaNLP Lab, Pisa, Italy. E-mail: \{alessio.miaschi,felice.dellorletta\}@ilc.cnr.it\\
\affmark[3] Universitat Politècnica de Catalunya, Department of Computer Science, Barcelona, Spain. E-mail: jordi.marco@upc.edu}
}

\date{Received: date / Accepted: date}

\maketitle

\begin{abstract}
Mobile app review analysis presents unique challenges due to the low quality, subjective bias, and noisy content of user-generated documents. Extracting features from these reviews is essential for tasks such as feature prioritization and sentiment analysis, but it remains a challenging task. Meanwhile, encoder-only models based on the Transformer architecture have shown promising results for classification and information extraction tasks for multiple software engineering processes. 
This study explores the hypothesis that encoder-only large language models can enhance feature extraction from mobile app reviews. 
By leveraging crowdsourced annotations from an industrial context, we redefine feature extraction as a supervised token classification task. Our approach includes extending the pre-training of these models with a large corpus of user reviews to improve contextual understanding and employing instance selection techniques to optimize model fine-tuning. Empirical evaluations demonstrate that these methods improve the precision and recall of extracted features and enhance performance efficiency. 
Key contributions include a novel approach to feature extraction, annotated datasets, extended pre-trained models, and an instance selection mechanism for cost-effective fine-tuning. This research provides practical methods and empirical evidence in applying large language models to natural language processing tasks within mobile app reviews, offering improved performance in feature extraction.
\end{abstract}

\section{Introduction}\label{sec:introduction}

Large language models (LLMs) have become a pervasive method for redefining cognitively challenging software engineering tasks based on the natural language processing (NLP) of textual documents~\citep{Hou2024}. 
As practitioners explore the potential of introducing these models into their day-to-day processes, multiple key design and evaluation factors are undermined or even neglected. This includes proper model analysis and selection~\citep{Perez202111054}, exploration of optimization mechanisms~\citep{Schick20212339}, explainability of results~\citep{Zini2022} and generalization of research outcomes~\citep{Jiang20202177}. Ignoring these dimensions can lead to low functional suitability, decreased performance efficiency, increased resource consumption and lack of potential for reusability and knowledge transfer.

In the context of Requirements Engineering (RE), effective adoption and proper use of LLMs have been elicited as key challenges for future work in the field~\citep{Frattini2024,Ronanki2023,Fantechi2023}. As a text-based document-driven community, both generated by developers (e.g., textual requirements, user stories, test cases) and by users (e.g., bug reports, software issues, app reviews), information extraction from these documents is key to support requirements elicitation~\citep{Ronanki2023}, defect repair~\citep{Ferrari20183684}, prioritization~\citep{Malgaonkar2022}, information extraction~\citep{Sleimi2021}, feedback analysis~\citep{Dalpiaz201955}, and release planning~\citep{McZara20151721,Sharma2019934}. Extracting descriptors, metadata, entities and categories from these documents allows practitioners to categorize large amounts of data and build analytical processes, which is especially useful for continuously analysing large amounts of user-generated documents~\citep{Li2018108,vanVliet2020143}. 
A particular example are opinion mining methods for mobile app reviews~\citep{Dabrowski2022}, for which three major tasks can be identified from the literature: (1) review classification (e.g., \textit{bug report}, \textit{feature request}~\citep{Maalej2015}); (2) sentiment analysis (e.g., \textit{positive}, \textit{negative}~\citep{ZHANG2014458}); and (3) app feature extraction (e.g., \textit{send message}, \textit{make video call}~\citep{Johann2017}). Concerning the latter, the automatic extraction of mobile app features (i.e., functions or quality characteristics of a mobile app from the user perspective) supports multiple feature-oriented decision-making tasks, including feature prioritization~\citep{Scalabrino201968} and feature-oriented sentiment analysis~\citep{Guzman2014}. 

While some feature extraction methods have been proposed by leveraging syntactic-based pattern matching techniques~\citep{Johann2017,Guzman2014,DRAGONI20191103}, several challenges remain~\citep{Dabrowski2022}. For starters, agreement towards what constitutes a \textit{feature} is typically low, as it is considered a particularly cognitively subjective task~\citep{DABROWSKI2023102181}. In addition, app reviews are low-quality user-generated documents, subjectively biased, grammatically incorrect, relatively short, filled with noisy content and even potentially generated by bots~\citep{Araujo2022}. Hence, syntactic-based and machine learning (ML) methods for feature extraction typically struggle especially in the context of app reviews, reporting an overall low recall and hence missing multiple feature mentions from a large corpus of reviews~\citep{DABROWSKI2023102181}.

To address these challenges, encoder-only LLMs, which rely solely on the encoder component of the Transformer architecture~\citep{Minaee2024}, have emerged as a promising solution. These models are particularly well-suited for tasks such as classification, named-entity recognition (NER), and information extraction. 
In this context, this study aims to validate the following hypothesis:


\begin{enumerate}[label=\textbf{H\(_\arabic*.\)}, leftmargin=*]
    \item Encoder-only LLMs can be leveraged to improve the state of the art of feature extraction in the context of mobile app reviews.
\end{enumerate}

To this end, our approach leverages crowdsourced user annotations from an industrial context to redefine the feature extraction task as a supervised token classification (e.g. NER) task. We compare the performance of multiple encoder-only LLMs with a classification layer on top to extract subsets of tokens referring to particular features for a given mobile app. 

Using this approach as a baseline, we complement our research with two additional hypotheses to explore the potential of improving the functional correctness and time behaviour of our LLM-based approach:

\begin{enumerate}[label=\textbf{H\(_\arabic*.\)}, start=2, leftmargin=*]
    \item Extending the pre-training of general-purpose LLMs (i.e., additional pre-training on domain-specific data) with a large corpus of user reviews can improve the representation of the mobile app user review context, improving the functional correctness of feature extraction.
    \item Instance selection of crowdsourced user reviews (i.e., filtering and selecting relevant instances for fine-tuning) can optimize model fine-tuning, improving the time behaviour of feature extraction while maintaining - or even improving - functional correctness.
\end{enumerate}





As a result, our research conveys the following contributions\footnote{Source code and datasets for replication of all experiments and full evaluation artefacts are available in the GitHub repository: \url{https://github.com/gessi-chatbots/t-frex}. The README file includes reference to models published on HuggingFace.}:

\begin{enumerate}[label=\textbf{C\(_\arabic*.\)}, leftmargin=*]
    \item A ground-truth dataset of 23,816 reviews from 468 apps belonging to 10 popular Google Play categories, annotated with 29,383 feature mentions.
    \item A proposal for automatically leveraging crowdsourced, user-generated feature annotations from an industrial context into mobile app reviews.
    \item A novel approach to redefining feature extraction from mobile app reviews as a token classification task.
    \item A curated dataset of 654,123 mobile app reviews from 832 apps belonging to 32 popular Google Play categories, used for extended pre-training.
    \item A collection of foundational encoder-only LLMs with extended pre-training for mobile app reviews NLP-based tasks.
    \item A document-based instance selection method for token classification tasks to support cost-effectiveness assessment of fine-tuning LLMs for NER.
    \item A collection of fine-tuned encoder-only LLMs for feature extraction, combining extended pre-training and instance selection mechanisms.
    \item An empirical analysis of the correctness and the execution time variations across the combination of (1) different types of encoder-only LLMs, (2) extended pre-training settings, and (3) instance selection data partitions.
\end{enumerate}

This research extends our previously published work~\citep{Motger2024saner}. Contributions C\(_1\) $\rightarrow$ C\(_3\) correspond to those already covered in the original publication, which we refer to as the T-FREX (\textit{\textbf{T}ransformer-based \textbf{F}eatu\textbf{R}e \textbf{EX}trac\-tion}) baseline design. To minimize overlap, we limit the scope of the empirical evaluation of the T-FREX baseline to relevant aspects necessary for self-containment and comparative evaluation in this publication. Additionally, we expand on the design and development details of our baseline approach, focusing on LLM-related topics (i.e., annotation transfer, model selection) that were not covered in detail in the original publication. Contributions C\(_4\) $\rightarrow$ C\(_8\) pertain exclusively to new contributions presented in this publication.

The structure of this paper is organized as follows. Section~\ref{sec:background} covers background literature and terminology, including feature extraction, token classification, extended pre-training, and instance selection. Section~\ref{sec:method} defines the research method, including the definition of the sample study. Section~\ref{sec:system-design} presents the T-FREX system design, including baseline and extended versions. Section~\ref{sec:evaluation} presents the evaluation design, the dataset and the experiment results. Section~\ref{sec:discussion} summarizes the discussion of the research questions. Section~\ref{sec:related-work} summarizes related work. Finally, Section~\ref{sec:conclusions} concludes our research.


\section{Background}\label{sec:background}

\subsection{Feature extraction}\label{sec:back-feat-ex}

A \textit{feature} is a specific function or capability within a mobile application that serves a particular purpose or fulfils a specific need~\citep{DABROWSKI2023102181}. Formal definitions typically refer to features either with software-related terminology (e.g., system capabilities, functional requirements~\citep{Wiegers2013}) or from a user-centered perspective (e.g., characteristics~\citep{KangFeatureOrientedDomain1990},  properties~\citep{Harman2012}). Beyond their formalization, features represent distinct characteristics of a given mobile app, designed to execute a clear task, comply with a particular qualitative expectation, or meet a specific user need. For instance, in the following review:

\begin{quote}
    \centering
    \textit{The feature for \textcolor{feature}{\textbf{sharing notes}} with other users is very handy.}
\end{quote}

The feature \textit{sharing notes} refers to a particular function which implies an actionable use case for a particular interaction between the user and the mobile app. On the other hand, in the following review:

\begin{quote}
    \centering
    \textit{This is the perfect \textcolor{feature}{\textbf{lightweight}} app for when you just want radar without all the other baloney bundled with it.}
\end{quote}

The feature \textit{lightweight} refers to a non-functional (i.e., quality) characteristic of the mobile app. Finally, in the following review:

\begin{quote}
    \centering
    \textit{I enjoy \textcolor{feature}{\textbf{meeting new people}} in my area and creating routes.}
\end{quote}

The feature \textit{meeting new people} refers to a particular capability that facilitates user engagement and community building within the app. 

\begin{figure*}[t]
\centerline{\includegraphics[width=\textwidth]{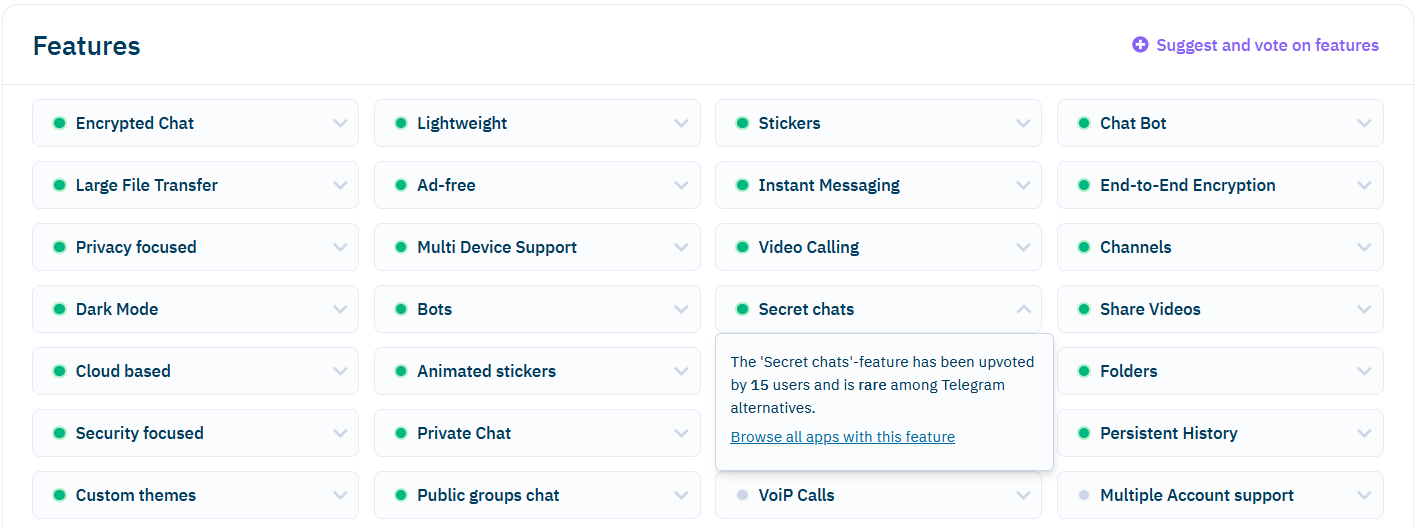}}
\vskip 6pt
\caption{List of Telegram features upvoted by users in AlternativeTo.}
\label{fig:features-telegram}
\end{figure*}

Features have become a core descriptor for app review mining activities~\citep{Dabrowski2022}. Furthermore, features are also used to support user-oriented services, including mobile app recommendation~\citep{Palomba2015291}, search-based algorithms~\citep{Chen201663} and personalization~\citep{Laranjo2021422}. Figure~\ref{fig:features-telegram} illustrates a practical example of features being used as categorization tags to support alternative software recommendations in AlternativeTo\footnote{Source: \url{https://alternativeto.net/software/telegram/about/}}. Features are used to cluster and identify similar mobile apps and provide recommendations based on similar features. Additionally, rare features are also highlighted to illustrate the app's singularity. However, these features are extracted based on manual user votes, limiting their generalization, and causing biased, unbalanced representativity and incomplete feature data.

\textit{Feature extraction} refers to the automatic identification of features mentioned within a textual document, including app descriptions and user reviews~\citep{Dabrowski2022}. While formal methods are mostly based on syntactic-based pattern-matching and topic modelling (see Section~\ref{sec:related-work}), several challenges remain. From a formalization perspective, scoping and stating what constitutes a feature is highly subjective, leading to discrepancies affecting methodological developments and even different application scenarios between industrial and research environments~\citep{Johann2017}. From a methodological perspective, due to the nature and authorship of reviews, these methods tend to struggle when processing noisy, grammatically inaccurate and complex documents~\citep{Shah2019}. From an evaluation point of view, this leads to low recall values, producing high false negatives and therefore limiting the added value of an automatic approach by missing multiple feature candidates~\citep{DABROWSKI2023102181}. 


\subsection{Token classification and NER}

Token classification is a fundamental task in the NLP field where the goal is to assign a particular label (i.e., class) to each token within a text sequence~\citep{Naveed2024}. Given a review $r$, let $T(r)$ denote the sequence of tokens in $r$. Formally, if $r$ consists of $n$ tokens, then $T(r) = [ t_1, t_2, \ldots, t_n ]$, where $t_i$ represents the $i$-th token in the sequence. Hence, token classification involves assigning one of the possible classes from a predefined set $C$ to each token $t_i$ in the sequence. As a result, the annotated sequence consists of pairs of tokens and their assigned classes, such as $[ (t_1, c_1), (t_2, c_2), \ldots, (t_n, c_n) ]$, where each $c_i$ represents the label assigned to the corresponding token $t_i$.



NER is a specialized form of token classification where the goal is to identify and classify specific entities within a text into predefined categories such as names of people, organizations, locations, dates, and other domain-specific entities~\citep{Hou2024}. This is achieved using the IOB tagging scheme, where each token in a sequence is assigned a label that indicates whether it is at the beginning ($B-$), inside ($I-$), or outside ($O$) of a named entity. 

In this research, we redefine the feature extraction task from mobile app reviews as a token classification task to identify and categorize features as named entities within a given review. In particular, following the structure of a NER task, we defined the set of possible classes $C$ as follows:
\begin{equation*}
C = \{ \text{\textit{B-feature}}, \text{\textit{I-feature}}, \text{\textit{O}} \}
\end{equation*}
Hence, given a review $r$ with tokens $T(r) = [ t_1, t_2, \ldots, t_n ]$, the function $\phi$ for 
token-based feature extraction is defined as:
\begin{equation*}
\phi(t_i) = \begin{cases} 
\text{B-feature} & \text{if } t_i \text{ is the beginning of a feature}, \\
\text{I-feature} & \text{if } t_i \text{ is inside a feature but not the beginning}, \\
\text{O} & \text{if } t_i \text{ is not part of any feature}.
\end{cases} 
\end{equation*}

For example, consider the review $r$ consisting of the tokens:
\begin{equation*}
T(r) = [\text{``To"}, \text{``do"}, \text{``list"}, \text{``function"},\text{``is"}, \text{``not"}, \text{``working"}]
\end{equation*}

If ``\textit{to-do list}" is identified as a feature, the annotated sequence would be:
\begin{align*}
[ &(\text{``To"}, \text{\textit{B-feature}}), (\text{``do"}, \text{\textit{I-feature}}), 
(\text{``list"}, \text{\textit{I-feature}}), \\
&(\text{``function"}, \text{\textit{O}}), (\text{``is"}, \text{\textit{O}}), 
(\text{``not"}, \text{\textit{O}}), (\text{``working"}, \text{\textit{O}}) ]
\end{align*}


\subsection{LLMs and extended pre-training}\label{sec:back-ep}

LLMs are pre-trained on large document corpora from various, multidisciplinary textual sources~\citep{Naveed2024}. These models use a combination of unsupervised and semi-supervised learning tasks, making them suitable for specific downstream tasks such as feature extraction. Using a large corpus of domain-specific documents (e.g., mobile app reviews), the pre-training phase of these models can be extended - also known as continual pre-training - to improve their performance on such tasks~\citep{Cagatay2024}. Several domains such as healthcare~\citep{Carrino2022193}, education~\citep{Liu2023666}, mathematics~\citep{Gong20225923}, or even less specialized domains such as product reviews~\citep{Jiang2023} have demonstrated the benefits of domain-specific extended pre-training. 
This results in enhanced contextual understanding, enabling models to have a finer-grained understanding of language nuances and context-specific knowledge. It also allows models to adapt to the specific vocabulary and stylistic elements of the domain, which may differ significantly from the data seen during the initial pre-training phases.

Extending LLM pre-training requires a large dataset within the target domain, ranging from 2 to 95 million tokens~\citep{Zixuan2023,Liu2023666,Jiang20202177,Carrino2022193}, depending on the domain specificity. Furthermore, it requires extensive computational resources due to the increased amount of data and the complexity of the training processes~\citep{Jiang20202177}. In addition, extending the pre-training entails some threats to validity, such as data bias and generalization to other tasks~\citep{Cagatay2024}.

Based on model selection (see Section~\ref{sec:system-design}), this paper focuses on two primary pre-training tasks: masked language modelling (MLM) and permutative language modelling (PLM). 
MLM involves masking a portion of the input tokens and training the model to predict the masked tokens based on the context provided by the unmasked tokens. For instance, given the review ``\textit{Sleep tracking is not working}", the model might be trained with the following masked token:

\begin{center}
\textit{Sleep \textcolor{orange}{\textbf{[MASK]}} is not working .} 
\end{center}

and learn to predict ``\textit{tracking}" as a suitable token in the context of mobile apps and features. 
On the other hand, PLM shuffles the order of the input tokens and trains the model to predict the original sequence. For example, from the same review, the model could be presented with:

\begin{center}
\textit{tracking Sleep is working not .}
\end{center}

and trained to reconstruct the original sequence, especially focusing on the original order of the ``\textit{sleep tracking}" feature.



\subsection{Instance selection}\label{sec:back-is}

Instance selection involves identifying and retaining the most relevant documents (e.g., mobile app reviews) from a corpus while filtering out non-relevant or redundant instances~\citep{Cunha2023}. This technique aims to enhance the efficiency and effectiveness of machine- and deep -learning models. Efficiency can be improved by reducing large datasets filtering non-relevant documents (i.e., documents not affecting the task performance in terms of functional correctness), especially when training on the entire corpus may be computationally prohibitive~\citep{Wilson2000}. Additionally, effectiveness can be improved by removing document instances that might be noisy, mislabeled, or highly redundant~\citep{Carbonera2017}. There are multiple instance selection strategies and methods according to the goal for filtering document instances. This includes removal of mislabeled~\citep{Wilson1972} or noisy~\citep{Wilson2000} documents using gold-standards, selection of highly representative documents through density measures~\citep{Carbonera2015}, and clustering-based approaches~\citep{MORAN2022794}.

Instance selection has been widely addressed in multiple NLP tasks such as text classification~\citep{Cunha2023}, sentiment analysis~\citep{Onan2016167}, text generation~\citep{Chang20218} and information extraction~\citep{Cardellino2015483}. However, few studies explore the potential of instance selection methods for NER tasks. And ultimately, these focus either on label transfer or propagation~\citep{Lu202139568} and random instance selection~\citep{Ferraro20241}.



\section{Study design}\label{sec:method}

\subsection{Goal and research questions}\label{sec:research-questions}

The main goal of this research is \textbf{to generate insights and empirical evidence about the use of encoder-only LLMs for feature extraction tasks in the context of mobile app reviews}. To this end, we elicited the following evaluation-oriented research questions~\citep{Shaw2003}:

\begin{enumerate}[label=\textbf{RQ\(_\arabic*.\)}, leftmargin=*]
    \item How effective are encoder-only LLMs at extracting features from mobile app reviews?\label{rq1}
    \item How does extending the pre-training of encoder-only LLMs improve the effectiveness of feature extraction from mobile app reviews?\label{rq2}
    \item How do instance selection methods improve the effectiveness of LLM-based feature extraction from mobile app reviews?\label{rq3}
    \item How does the combination of extended pre-training and instance selection improve the effectiveness of LLM-based feature extraction from mobile app reviews?\label{rq4}
\end{enumerate}

RQ\(_1\) addresses the validation of H\(_1\). RQ\(_2\) pertains to the validation of H\(_2\). Finally, RQ\(_3\) and RQ\(_4\) focus on the validation of H\(_3\).

\subsection{Research method}\label{sec:research-method}

\begin{figure*}[h]
\centerline{\includegraphics[width=\textwidth]{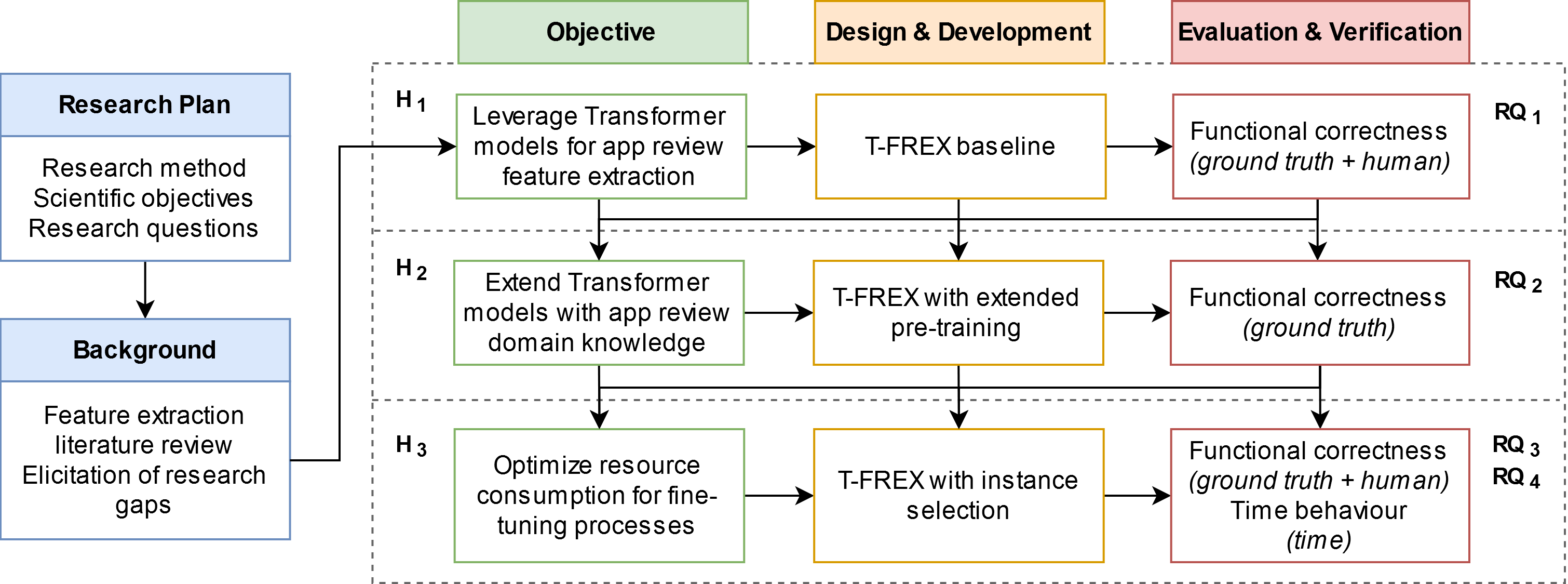}}
\vskip 6pt
\caption{Research method}
\label{fig:research-method}
\end{figure*}

Figure~\ref{fig:research-method} illustrates a general overview of the research method conducted in this study.
Following a Design Science (DS) methodology for software engineering research~\citep{Engström20202630}, we refined the main goal of this research into three scientific sub-objectives focusing on the validation of H\(_1\) $\rightarrow$ H\(_3\). We aligned each Design Science iteration with a particular sub-objective, leading to three iterations covering (\textit{i}) objective refinement, (\textit{ii}) design and development of the solution, and (\textit{iii}) empirical evaluation and verification of results. Concerning evaluation, we use the ISO/IEC 25010 software product quality model~\citep{iso25010} to focus on two quality characteristics for the evaluation: \textit{functional correctness} 
with respect to ground truth data and human evaluation; and \textit{time behaviour} of fine-tuning processes. 
Extended details of the evaluation design are depicted in Section~\ref{sec:evaluation}.

\subsection{Sample study: Google Play and AlternativeTo}

We shaped our research as a sample study using data extracted from the Google Play App Store\footnote{https://play.google.com/store/apps} to minimize obtrusiveness and maximize generalization of our research findings~\citep{Stol2018}. Minimum obtrusiveness is achieved by limiting manipulation of the research settings to data collection from existing repositories (i.e., app stores), constraining instrumentation to NLP-based pre-processing (see Section~\ref{sec:data-pre-processing}). Maximum generalization is aimed by focusing on cross-domain documents in the context of mobile app reviews, using data from different mobile app categories, focusing on popular data sources, and designing evaluation settings exploring the generalization of research findings to unknown domains (see Section~\ref{sec:evaluation-baseline}).

Google Play is the largest app store worldwide, hosting more than 3.5 million apps, followed by the Apple App Store with 1.6 million apps~\citep{Statista2024}. Its market quote reaches 2,500 million users worldwide~\citep{google_play_works_2024}, offering a large catalogue of mobile apps from 32 app categories~\citep{google_play_support_2024}, excluding games. The potential of app stores for empirical software engineering research has not been overlooked, leading to multiple studies benefitting from the available data, including reviews, ratings, app descriptions and changelogs, among others~\citep{McIlroy20161067,McIlroy20161346,Hassan20181275}. In this research, we focus on the collection of mobile app reviews from popular Google Play categories to evaluate the performance of the feature extraction process (see Section~\ref{sec:dataset} for details on the collected dataset).

In addition to Google Play, we complement our data collection with crowdsourced feature annotations generated by users from AlternativeTo\footnote{https://alternativeto.net/}, a software recommendation platform for finding alternatives for a given software product, including web-based, desktop and mobile apps. AlternativeTo has a catalogue of 120,000 apps and has collected feedback from almost 2 million users worldwide. As illustrated in Figure~\ref{fig:features-telegram} (see Section~\ref{sec:back-feat-ex}), users can upvote features pertaining to a particular software. These features are then used for various use cases, such as measuring similarity with alternative apps, looking for apps with a given feature and measuring feature frequency to determine its popularity or rareness. We extract and use these annotations as ground truth for the empirical evaluation of our approach. By leveraging crowdsourced user annotations, we argue that our approach aligns with the concept of features being used in a real context (i.e., minimizing manipulation to data collection).

Details on the data collection process are presented in Section~\ref{sec:system-design}. A summarized overview of the collected dataset of reviews and features is presented in Section~\ref{sec:evaluation}.


\section{System design}\label{sec:system-design}

\subsection{T-FREX baseline}\label{sec:baseline-fine-tuning}

\begin{figure*}[t]
\centerline{\includegraphics[width=\textwidth]{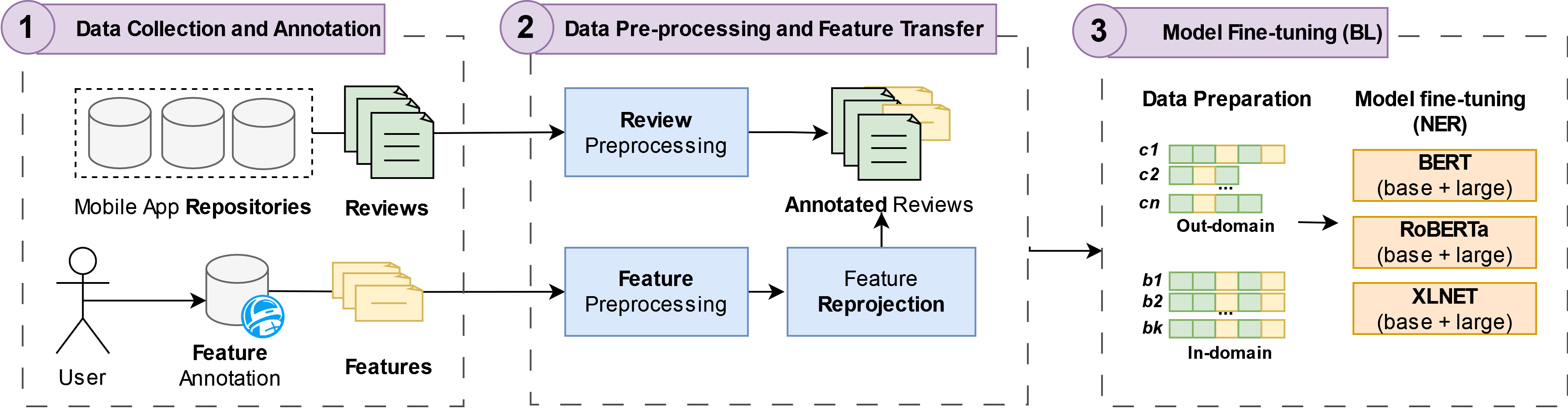}}
\vskip 6pt
\caption{Design of T-FREX baseline}
\label{fig:design-BL}
\end{figure*}

Figure~\ref{fig:design-BL} illustrates a summarized overview of the T-FREX baseline design proposal. T-FREX baseline is composed of three main stages: \circled{1} data collection and annotation of user reviews and features; \circled{2} data pre-processing and feature transfer from apps to reviews; and \circled{3} model selection and fine-tuning for the token classification function $\phi$.

\subsubsection{Data collection and annotation}\label{sec:data-pre-processing}

Data collection \circled{1} is composed of two data artefacts: reviews and features.

\begin{itemize}
    \item \textbf{Reviews dataset.} We built on our previous work~\citep{Motger2023} by reusing a dataset of 639 mobile apps with 622,370 reviews from Google Play and other Android mobile app repositories belonging to multiple, heterogeneous mobile app categories (e.g., communication, health and fitness, maps and navigation, lifestyle...). We applied the following modifications to this dataset: (1) we focused on categories exceeding a minimum threshold for statistical representativeness (i.e., including $\geq 5$ apps); and (2) we removed all categories related to game apps, as the concept of \textit{feature} in the scope of this research does not apply to the concept of \textit{feature} in game mobile apps~\citep{DABROWSKI2023102181}. This process resulted in 364,220 reviews from 468 mobile apps from 10 popular Google Play categories.
    \item \textbf{Features dataset.} Using the set of reviews, we implemented web scraping mechanisms to access archived cached versions from AlternativeTo using the Wayback Machine Internet Archive project\footnote{http://web.archive.org/} to automatically extract feature annotations by users. Using the BeautifulSoup library\footnote{https://pypi.org/project/beautifulsoup4/}, we dynamically inspect the AlternativeTo profile page of each app of the reviews dataset, and we locate and parse the `Features' section (as shown in Figure~\ref{fig:features-telegram}) to extract a flat list of features for each app (e.g., for Telegram, we collect features such as \textit{`Encrypted Chat'}, \textit{`Large File Transfer'} and \textit{`Animated stickers'}). This resulted in 198 distinct features for 468 apps.
\end{itemize}

Extended details on the reviews and features dataset, including a category-oriented analysis, are presented in Section~\ref{sec:dataset}.

\subsubsection{Data pre-processing and feature transfer}\label{sec:annotation-transfer}

\begin{algorithm}[t]
\caption{Feature Annotation Transfer}
\label{algorithm:feature-transfer}
\begin{algorithmic}[1]
\Require $F = \{f_1, f_2, \ldots, f_q\}$ \Comment{Set of crowdsourced feature annotations}
\Require $R = \{r_1, r_2, \ldots, r_m\}$ \Comment{Set of app reviews}
\Require $L = \{O, B\text{-}feature, I\text{-}feature\}$ \Comment{Name entity labels}
\Ensure $R$ annotated with labels from $L$

\State \textbf{Pre-process} $F$ and $R$ using Stanza's neural pipeline:
\State \hspace{1em} $F\textsuperscript{$\prime$} \gets \text{PreProcess}(F)$
\State \hspace{1em} $R\textsuperscript{$\prime$} \gets \text{PreProcess}(R)$

\For{each $r \in R\textsuperscript{$\prime$}$}
    \For{each $t \in T(r)$}
        \State $t.label \gets O$ \Comment{Initialize with default label (i.e., non-feature token)}
    \EndFor
    \State $app_r \gets r.app$ \Comment{Get the app which $r$ belongs to}
    \For{each $f \in F\textsuperscript{$\prime$}$}
        \State $app_f \gets f.app$ \Comment{Get the app where $f$ was annotated}
        \If{$app_f = app_r$}
            \If{$T(f) \subseteq T(r)$}
                \State $start \gets \text{index of first token in } T(f) \text{ within } T(r)$
                \State $end \gets start + |T(f)| - 1$
                \State $T(r)[start].label \gets B\text{-}feature$ \Comment{Annotate beginning feature tokens}
                \For{$i \gets start + 1$ \textbf{to} $end$}
                    \State $T(r)[i].label \gets I\text{-}feature$ \Comment{Annotate internal feature tokens}
                \EndFor
            \EndIf
        \EndIf
    \EndFor
\EndFor

\State \textbf{Output} the annotated corpus $R$ in CoNLL-U format

\end{algorithmic}
\end{algorithm}

Algorithm~\ref{algorithm:feature-transfer} depicts in detail the steps for transferring the features collected from AlternativeTo into annotations within the dataset of reviews. For review and feature pre-processing \circled{2}, we apply a common natural language pre-processing stage using Stanza's neural pipeline\footnote{https://stanfordnlp.github.io/stanza/neural\_pipeline.html} for syntactic annotation and entity extraction from both the reviews and features datasets (lines 1-3). The pipeline was composed of the following steps: (\textit{i})~tokenization, (\textit{ii}) multi-word token expansion, (\textit{iii}) PoS tagging, (\textit{iv}) morphological feature extraction, and (\textit{v}) lemmatization. The output of this pipeline is formatted using the CoNLL-U format\footnote{https://universaldependencies.org/format.html}. After the pre-processing step, the corpus is ready for the annotation process. Each token $t \in T(r), \forall r \in R$ is initialized with the default label $O$ (lines 5-7). The algorithm looks for all matches between feature tokens and review tokens (lines 9-21). If a match is found for a particular review $r$ and feature $f$ for a given mobile app $app$ (lines 11-12), which means that users from AlternativeTo voted $f$ as a feature from $app$, then each token resulted from the intersection $T(r) \cap T(f)$ is annotated with \textit{B-feature} or \textit{I-feature} according to the position of the token within the original feature (lines 13-18). 

This process resulted in 29,383 feature annotations over 23,816 app reviews. Section~\ref{sec:dataset} provides extended details on the resulting feature annotations after the feature transfer process.



\subsubsection{Model fine-tuning}\label{sec:model-fine-tuning}

Stemming from recent literature reviews in the field of LLMs~\citep{Hou2024,Naveed2024,Zhao2023}, we compared different encoder-only LLMs suitable for our evaluation and comparative analysis. We focused on encoder-only architecture due to their inherent suitability for classification tasks~\citep{Hou2024}. In addition, we also excluded decoder-only models (also known as generative models) due to their size and resource consumption. These models present limited applicability in large-scale contexts such as user review mining, especially in terms of memory, computational resources and time constraints. Particularly, in this study, we selected the following models:

\begin{itemize}
    \item \textbf{BERT}, considered the first encoder-only LLM, is renowned for its advanced contextual understanding due to its bidirectional nature~\citep{devlin2019bert}. It is pre-trained using the MLM objective, which enhances its ability to grasp context from both directions, making it effective for token-level tasks such as NER~\citep{Broscheit2019}. For these reasons, we use BERT as a baseline LLM for NER tasks.
    \item \textbf{RoBERTa} improves upon BERT's design and training methods through extended pre-training on a larger dataset with additional data, resulting in stronger language representations~\citep{liu2019roberta}. It also uses MLM for pre-training but outperforms BERT in many cases~\citep{liu2019roberta}. We include RoBERTa in our model evaluation due to its enhanced performance over BERT.
    \item \textbf{XLNet} uses a unique approach by combining autoregressive and bidirectional training, considering all possible word permutations during pre-training~\citep{yang2019xlnet}. This improves its ability to understand context and model token dependencies more effectively than traditional models. Unlike BERT and RoBERTa, XLNet employs a PLM training objective. Consequently, token dependencies are modelled differently. We evaluate XLNet's performance to compare its innovative training method against the MLM objectives of BERT and RoBERTa.
\end{itemize}

Encoder-only models have not significantly evolved over the past few years. As a result, while generative (decoder-only) models have experienced increased growth and extended research, models like BERT, RoBERTa, and XLNet still represent the state-of-the-art for encoder-only LLMs, offering robust performance across diverse tasks~\citep{Yang2023}. However, their validity in many specific scenarios still needs thorough assessment~\citep{Hou2024}.

Table~\ref{tab:model-comparative} provides the full lists of models used in this research, as well as some size-related features. For each model, we use both base and large versions. 

\begin{table}[]
\centering
\caption{Model features and fine-tuning parameters}
\resizebox{\columnwidth}{!}{%
\begin{tabular}{@{}lrrrrrr@{}}
\toprule
\textbf{model} & \textbf{data} &  \textbf{parameters} & \textbf{task} & \textbf{epochs} & \textbf{learning rate} & \textbf{batch size} \\ \midrule
\textbf{BERT\textsubscript{base}}  & 16 GB  & 110 M & MLM & 2 & 2e-5 & 16 \\
\textbf{BERT\textsubscript{large}} & 16 GB  & 336 M & MLM &2 & 2e-5 & 16 \\
\textbf{RoBERTa\textsubscript{base}}       & 160 GB & 125 M & MLM &2 & 2e-5 & 16 \\
\textbf{RoBERTa\textsubscript{large}}      & 160 GB & 355 M & MLM &2 & 2e-5 & 8  \\
\textbf{XLNet\textsubscript{base}}   & ~16 GB  & 110 M & PLM & 2 & 3e-5 & 16 \\
\textbf{XLNet\textsubscript{large}}  & ~113 GB & 340 M & PLM & 2 & 3e-5 & 8  \\ \bottomrule
\end{tabular}%
}
\label{tab:model-comparative}
\end{table}



After model selection, the corpus of reviews $R$ is divided into $k$ subsets $\{R_1, R_2, \ldots, R_k\}$. For each fold $j \in \{1, 2, \ldots, k\}$, a fine-tuning iteration involves using the subset $R_j$ as the test set, while the remaining subsets $R \setminus R_j$ are combined to form the training set. This process is repeated $k$ times, with each subset $R_j$ used exactly once as the test set. We design two different strategies for data preparation:

\begin{itemize}
    \item \textbf{In-domain learning}. The set of reviews $R$ is split into $k$ partitions, each containing the same proportion of reviews from each mobile app category as the original dataset. This ensures a balanced representation of each category in each partition. The in-domain learning setting evaluates the model's performance when it is trained on data from all domains, ensuring a diverse and representative training set.
    
    \item \textbf{Out-of-domain learning}. The data is split into $k$ partitions, where $k$ is the number of different mobile app categories. Each partition contains reviews exclusively from one category. This setup evaluates the model's performance in extracting features from a domain it was not trained on, testing its generalizability to new, unseen categories.
\end{itemize}

After model selection and data preparation, we design the fine-tuning process \circled{3} as follows:

\begin{enumerate}
    \item \textbf{Data processing}. Loading of train and test datasets, transformation from CoNLL-U format to a dataset compatible with the HuggingFace datasets library, and tokenization of user reviews according to the model architecture used in each evaluation sequence. For BERT, we use WordPiece tokenizer, while for RoBERTa and XLNet we use SentencePiece tokenizer. The main difference involves the management of special tokens (e.g., BERT uses [CLS] classification token) and tokenization granularity (e.g., RoBERTa and XLNet employ more fine-grained tokenization where multiple tokens can belong to the same word). 
    \item \textbf{Model loading}. Loading the model from the model library. For T-FREX baseline, checkpoints are directly loaded using the HuggingFace model library API. This step involves initializing the model architecture and loading original pre-trained weights to leverage transfer learning. 
    \item \textbf{Training setting}. Configuring the training parameters, including number of epochs, learning rate and batch size. Table~\ref{tab:model-comparative} reports experimentation details used in this study for each model. Variations relate to limitations of computational resources, assessing the balance between memory usage and performance efficiency. This configuration is used in all research iterations, including T-FREX with extended pre-training (Section~\ref{sec:extended-pre-training}) and with instance selection (Section~\ref{sec:instance-selection})
    \item \textbf{Training}. Fine-tuning process of the proper model (i.e., BERT, RoBERTa, XLNet) with the training set. This step involves iterative optimization of model parameters using backpropagation and gradient descent. The training process aims to minimize the loss function, improving the model's ability to predict feature tokens accurately. Regular monitoring of training metrics, such as loss and accuracy, is conducted to ensure the model is learning effectively and to prevent overfitting. 
    \item \textbf{Evaluation}. Running model for inference to evaluate the performance with the test set.  This includes the collection of metrics used for evaluation in this study, such as precision, recall, and f-measures (see Section~\ref{sec:evaluation-design}). The evaluation process assesses the model's generalization capability and its effectiveness in extracting features from unseen user reviews and features. 
\end{enumerate}

Evaluation results for T-FREX baseline (RQ\(_1\)) are reported in Section~\ref{sec:evaluation-baseline}.

\subsection{T-FREX with extended pre-training}\label{sec:extended-pre-training}

\begin{figure*}[t]
\centerline{\includegraphics[width=\textwidth]{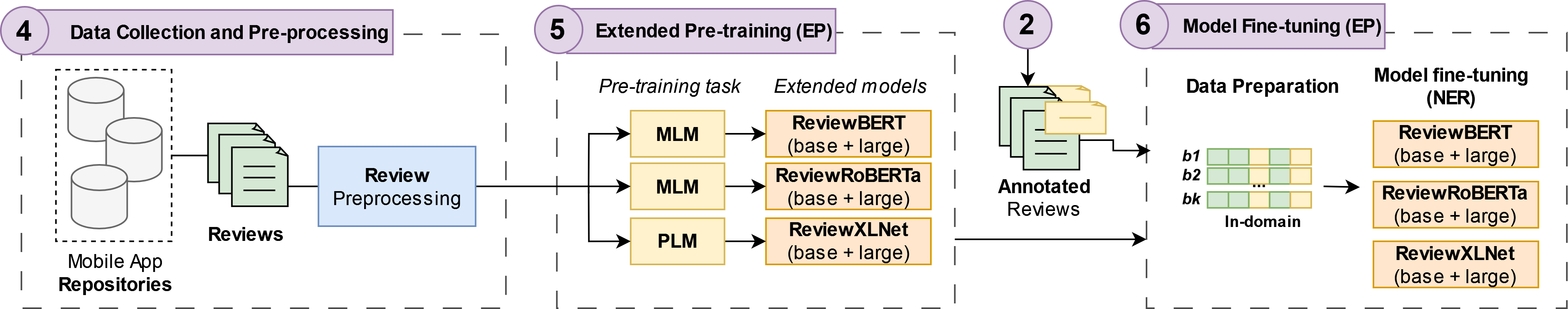}}
\vskip 6pt
\caption{Design of T-FREX with extended pre-training}
\label{fig:design-EP}
\end{figure*}

Figure~\ref{fig:design-EP} illustrates a summarized overview of the T-FREX design proposal with extended pre-training of LLMs. This approach is composed of three main stages: \circled{4} data collection and pre-processing; \circled{5} extension of the pre-training task of each LLM used in this research; and \circled{6} fine-tuning with extended models using the annotated ground truth generated during T-FREX baseline~\circled{2}.

\subsubsection{Data collection}

We build on our previous work in the generation of a dataset \circled{4} of mobile app reviews extended from the dataset used for T-FREX baseline~\citep{Motger2024}. This dataset consisted of 13,478,744 user reviews from 832 mobile apps belonging to 46 categories. To minimize obtrusiveness and guarantee minimal manipulation of the dataset and the context, we limited the pre-processing of such reviews to a minimal sanitization pipeline consisting of the following steps: (\textit{i}) converting text to UTF-8 encoding; (\textit{ii}) filter non-English reviews; and (\textit{iii}) remove duplicate text. Data is saved in CoNLL format for consistency with T-FREX baseline. In addition, to minimize computational consumption, we limited the dataset to an acceptable minimum size for extending pre-training using references of related work in the field of domain-specific extended pre-training (see Section~\ref{sec:back-ep}).
As BERT, RoBERTa and XLNet are pre-trained using token level tasks (i.e., MLM and PLM), we use the length of the corpus in terms of tokens as a metric to limit the dataset. Consequently, we reduced the dataset to 8,232,362 tokens pertaining to 622,352 reviews. 

\subsubsection{Extending models pre-training}

Based on model selection for T-FREX baseline (see Section~\ref{sec:model-fine-tuning}), we used the extended dataset to extend the pre-training \circled{5} of BERT, RoBERTa (with MLM) and XLNet (with PLM). Initially, CoNLL formatted data is converted into a HuggingFace dataset. The dataset is split into training and evaluation sets, followed by tokenization using a tokenizer specific to the model type (i.e., WordPiece for BERT, SentencePiece for RoBERTa and XLNet). The tokenized reviews are grouped into blocks of 128 tokens to ensure efficient training. Then, we prepare the fine-tuning process for either MLM or PLM tasks. This fine-tuning process is focused on reducing the value of the evaluation loss after each epoch, which we monitor and report during the evaluation process (see Section~\ref{sec:evaluation}). Training arguments are set using the same values as in Table~\ref{tab:model-comparative}, with the exception of the number of epochs, which we extended to 10 in order to analyse the evolution of the continual pre-training effect after several iterations. Hence, we save a model checkpoint after each epoch for future evaluation with token classification fine-tuning. 

\subsubsection{Model fine-tuning}

For fine-tuning with extended models \circled{6}, we repeated steps $1 \rightarrow 5$ from the fine-tuning process as defined for T-FREX baseline (see Section~\ref{sec:model-fine-tuning}) for each checkpoint and model saved during the extended pre-training stage \circled{5}. We used the set of annotated reviews from the T-FREX baseline design \circled{2}, and we limited the data preparation of T-FREX with extended pre-training to the in-domain data analysis. This is motivated by three reasons. First, the mobile app domain is a highly stable environment in terms of emerging mobile app categories, making in-domain learning the most common scenario, as variability in the list of mobile app categories is very limited. Second, the purpose of the out-of-domain analysis is to test the T-FREX baseline in a limiting, challenging scenario, exploring the strengths and weaknesses of a NER-based approach under unexpected circumstances (i.e., generalization to a new app category). Finally, extended pre-training and fine-tuning processes are computationally expensive, requiring high energy consumption. Specifically, the in-domain analysis itself entails a total of 60 fine-tuning processes (10 checkpoints $ \times$ 6 LLM instances). Hence, in addition to previous considerations and to promote sustainability, we limit the scope of our research to the most common scenario in the context of mobile apps (i.e., in-domain learning). 

Evaluation results for T-FREX with extended pre-training (RQ\(_2\)) are reported in Section~\ref{sec:eval-extended-pretraining}.

\subsection{T-FREX with instance selection}\label{sec:instance-selection}

\begin{figure*}[t]
\centerline{\includegraphics[width=\textwidth]{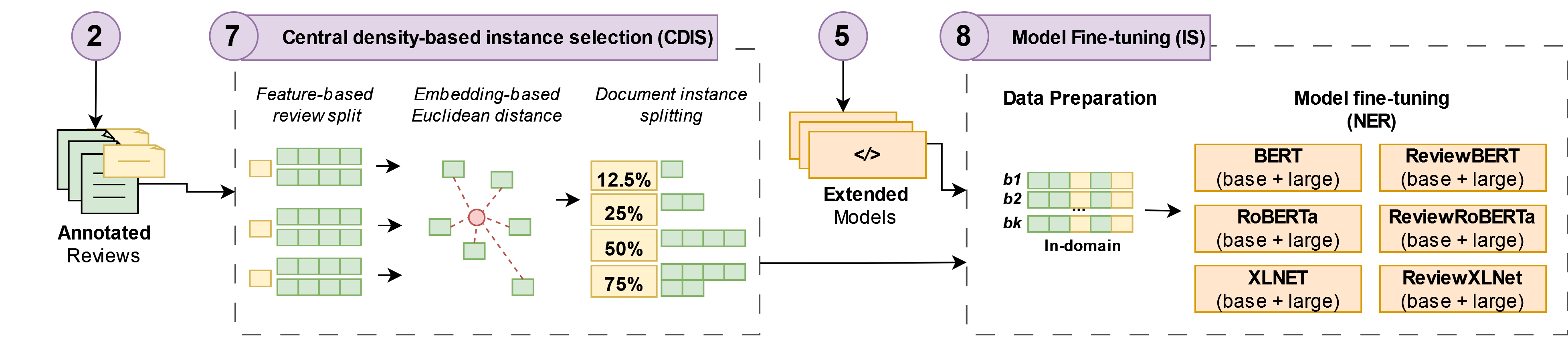}}
\vskip 6pt
\caption{Design of T-FREX with instance selection}
\label{fig:design-IS}
\end{figure*}

Figure~\ref{fig:design-IS} illustrates the T-FREX design proposal with instance selection of reviews. T-FREX with instance selection is composed of two main stages: \circled{7} instance selection of user reviews using a density-based instance selection mechanism; and \circled{8} model fine-tuning with original and extended LLM instances using the annotated ground truth generated during T-FREX baseline.

\subsubsection{Central density-based instance selection}

We propose an adapted version of a central density-based instance selection (CDIS) algorithm for classification tasks~\citep{Carbonera2016}. This approach \circled{7} focuses on redundancy removal for optimal resource consumption and increased accuracy in prediction quality. Specifically, our approach is focused on two main adaptations: (1) reshaping the instance selection criteria for a NER task (i.e., using different feature entities as criteria for document splitting); and (2) leveraging LLMs to generate contextualized embeddings for each document, compute centroids for a semantic space (i.e., reviews mentioning a given feature) and compute distances between documents (i.e., reviews) and the theoretical centroid.

This process is summarized in Algorithm~\ref{algorithm:is}. The algorithm takes the corpus of app reviews $R$ with annotated features $F$ as inputs \circled{2}. For each review $r \in R$, we generate its embedding using \text{BERT}\textsubscript{base}, which we select as a baseline representative of encoder-only LLMs. Then, we collect all reviews containing each feature $f \in F$ into corresponding sets $D[f]$. For each feature $f$, the algorithm aggregates the embeddings of the reviews mentioning $f$ and computes the centroid $C_f$ of these embeddings. The distance between each review embedding and the centroid $C_f$ is calculated using the Euclidean distance. Reviews are then sorted based on their distance to the centroid in descending order. Finally, the algorithm partitions the sorted reviews into four subsets based on the specified training data distributions: 12.5\%, 25\%, 50\%, and 75\%. The merged partitions for all features are then output as the result. Consequently, each partition maximizes semantic representativeness of reviews at feature level. This optimizes training sets by minimizing highly similar reviews which might impact negatively the fine-tuning process both from the functional correctness (i.e., overfitting or unbalanced semantic representation) and from the performance efficiency (i.e., unnecessarily large datasets) points of view.

\begin{algorithm}[h]
\caption{Instance Selection}
\label{algorithm:is}
\begin{algorithmic}[1]
\Require $R = \{r_1, r_2, \ldots, r_m\}$ \Comment{Corpus of app reviews with annotated features}
\Require $F = \{f_1, f_2, \ldots, f_q\}$ \Comment{Set of features}
\Require $BERT\textsubscript{base}$ \Comment{Pre-trained BERT base model}
\Ensure Review subset partitions based on training data distributions 

\For{each $r \in R$}
    \State $r.\text{embedding} \gets \text{BERT}.\text{compute\_embedding}(r.\text{raw\_text})$
\EndFor

\State $D \gets \{\}$
\For{each $f \in F$}
    \State $D[f] \gets \{r \in R \mid T(f) \subseteq T(r)\}$ \Comment{Collect reviews $r$ containing $f$}
\EndFor

\For{each $f \in F$}
    \State $E_f \gets \{\}$
    \For{each $r \in D[f]$}
        \State $E_f \gets E_f \cup \{r.\text{embedding}\}$ \Comment{Collect review embeddings containing $f$}
    \EndFor
    \State $C_f \gets \text{compute\_centroid}(E_f)$ \Comment{Compute centroid of embeddings}

        \State $r.\text{distance} \gets \text{euclidean\_distance}(r.\text{embedding}, C_f)$

    \State $D[f] \gets \text{sort}(D[f], \text{key} = r.\text{distance}, \text{reverse} = \text{True})$
\EndFor

\State $P \gets \{\text{0.125} \to \{\}, \text{0.25} \to \{\}, \text{0.50} \to \{\}, \text{0.75} \to \{\}\}$

\For{each $f \in F$}
    \State $n \gets |D[f]|$ \Comment{Total number of reviews containing $f$}
    \State $P[\text{0.125}] \gets P[\text{0.125}] \cup D[f][0 : \lceil 0.125 \cdot n \rceil]$
    \State $P[\text{0.25}] \gets P[\text{0.25}] \cup D[f][0 : \lceil 0.25 \cdot n \rceil]$
    \State $P[\text{0.50}] \gets P[\text{0.50}] \cup D[f][0 : \lceil 0.5 \cdot n \rceil]$
    \State $P[\text{0.75}] \gets P[\text{0.75}] \cup D[f][0 : \lceil 0.75 \cdot n \rceil]$
\EndFor

\State \textbf{Output} $P$ \Comment{The merged partitions for all features}
\end{algorithmic}
\end{algorithm}

\subsubsection{Model fine-tuning}

Similarly to the extended pre-training approach, we repeated steps $1 \rightarrow 5$ from the fine-tuning process as defined for T-FREX baseline with instance selection training splits \circled{8}. To explore the impact of instance selection in isolation (RQ\(_3\)) and in combination with extended pre-training (RQ\(_4\)), we fine-tune all data partitions with both the original and the extended checkpoints generated during the T-FREX fine-tuning with extended pre-training \circled{5}. This leads to a total of 264 fine-tuning processes (4 training sets $\times$ 6 model instances $\times$ 1+10 model checkpoints). Using the same criteria as in the extended pre-training, we limit the scope of analysis to the in-domain data preparation scenario.

Evaluation results for T-FREX with instance selection, both in isolation (RQ\(_3\)) and combined with extended pre-training (RQ\(_4\)), are reported in Section~\ref{sec:eval-instance-selection} and Section~\ref{sec:eval-combined}, respectively.


\section{Evaluation}\label{sec:evaluation}


\subsection{Design}\label{sec:evaluation-design}


\subsubsection{Functional correctness - Ground truth}\label{sec:eval-design-fs-gt}



For \textit{functional correctness}, we use collected reviews from Google Play and features from AlternativeTo as ground truth by assessing the correctness of $\phi(t_i)$ predictions $\forall t_i \in T(r)$ and $\forall r \in R$, where $R$ is the corpus of reviews used for evaluation (see Section~\ref{sec:dataset}). 
Based on model predictions, we focus on the following metrics for measuring functional correctness: precision, recall and f-measure. 

\begin{itemize}
    \item \textbf{Precision} ($p$) evaluates the proportion of correctly predicted feature-related tokens (\(c_i \in \{\text{\textit{B-feature}}, \text{\textit{I-feature}}\}\)) relative to all tokens predicted as feature-related, measuring how accurate the model is in identifying features.
    \item \textbf{Recall} ($r$) measures the proportion of correctly predicted feature-related tokens compared to all feature-related tokens in the ground truth, highlighting how many features were successfully captured by the model.
    \item \textbf{F-measure} ($f$) provides a balanced evaluation by calculating the harmonic mean of precision and recall, ensuring that both false positives and missed features are taken into account.
\end{itemize}





In the context of evaluating NLP-based RE tasks, app review analysis is identified as an example of a \textit{hairy} RE task. \textit{Hairy} tasks are defined as document-driven, non-algorithmic, manageable by experts on a small scale and unmanageable on a large scale~\citep{Berry2021}. Consequently, achieving high recall is key due to the need for close to 100\% recall in identifying all relevant answers. While precision remains important, we focus on maximizing recall to minimize overlooked feature mentions. Consequently, we propose using a different weighting for the f-measure to reflect the greater importance of recall over precision. In addition to $f_1$, as proposed by~\cite{Berry2021}, we also measure:

\begin{equation*}
    f_{\beta} = (1 + \beta^2) \cdot \frac{p \cdot r}{\beta^2 \cdot p + r}, \quad \beta = A_T / A_t
\end{equation*}

where $A_T$ is the average time to find a relevant feature when performing feature extraction manually, and $A_t$ is the average time to determine the validity of a potential feature. Hence, $\beta$ is the relative importance of recall (finding all relevant features) to precision (determining the validity of features) based on the relative cost between $A_T$ and $A_t$. Computation of $A_T$ and $A_t$ is done during the human validation process (see Section~\ref{sec:eval-design-fs-human-eval}).

For the in-domain setting, we measure these metrics using a $k$-fold cross-validation analysis with $k=10$ and report average values. For the out-of-domain setting, $k$ is determined by the number of mobile app categories. 

We exclude accuracy from the evaluation because we cannot control the exhaustivity of the annotations. Without certainty that all features of a given mobile app are annotated in AlternativeTo, we cannot ensure that tokens labelled by default as non-feature entities ($c_i = O$) in Algorithm~\ref{algorithm:feature-transfer} are correct.

\subsubsection{Functional correctness - Human evaluation}\label{sec:eval-design-fs-human-eval}

Crowdsourced features from AlternativeTo impose some limitations due to the lack of control of the annotation process (see Section~\ref{sec:ttv}). As some features might be overlooked, tokens labelled with $c_i \in \{\text{\textit{B-feature}}, \text{\textit{I-feature}}\}$ annotated with \textit{O} might be falsely detected as an FP instance. This limits the reliability of the precision metric. To address this, we incorporated an external human evaluation process strictly on newly predicted features, i.e., those not annotated in the original ground truth set. 
Building on this and the considerations raised in Section~\ref{sec:eval-design-fs-gt}, human evaluation goals include to (1) mitigate the limitation of our dataset's lack of exhaustivity, and (2) compute the $\beta$ to report a task-specific measure for correctness. 
This human evaluation process consisted of the following steps:

\begin{enumerate}
    \item \textbf{Questionnaire design.} We designed questionnaires for human assessment using QuestBase\footnote{https://questbase.com/} which were distributed to external annotators with Prolific\footnote{https://www.prolific.com/}. Specifically, we designed two different sets of questionnaires:
    \begin{enumerate}
        \item \textbf{Assessment for automatic feature extraction ($Q_{A}$)}. Figure~\ref{fig:eval-design-fp} illustrates a snapshot of the questionnaire to assess the validity of a feature automatically extracted by T-FREX. Each question presents to the annotator: (1) the app name, including a link to Google Play; (2) the app category; (3) the text of the review; (4) the proposed feature; and (5) the question. Annotators can confirm the feature proposal (\textit{Yes}), reject it (\textit{No}) or report it as not clear (\textit{I don't know}). We conducted iterative internal annotations to adjust the size (100 reviews), required time (15') and economic retribution ($2\pounds$). Each $Q_{A}$ questionnaire includes 5 control questions to reject annotators not passing a minimum performance requirement (i.e., 4/5 correct feature annotations with trivial examples from the ground truth).
        \item \textbf{Assessment for manual feature extraction (\textit{$Q_{M}$})}. Figure~\ref{fig:eval-design-mfe} illustrates a snapshot of the questionnaire to manually extract features from a given review. Annotators are presented with the same information as in previous questionnaires, with the exception of the question and the answers, which in this case is a free-text area. We reduced the size of each task (25 reviews) while keeping time (15') and retribution ($2\pounds$). $Q_{M}$ questionnaires include 3 control questions to reject annotators using a performance threshold (i.e., 2/3 correct feature annotations).
    \end{enumerate}

    \begin{figure*}[t]
    \centerline{\includegraphics[width=\textwidth]{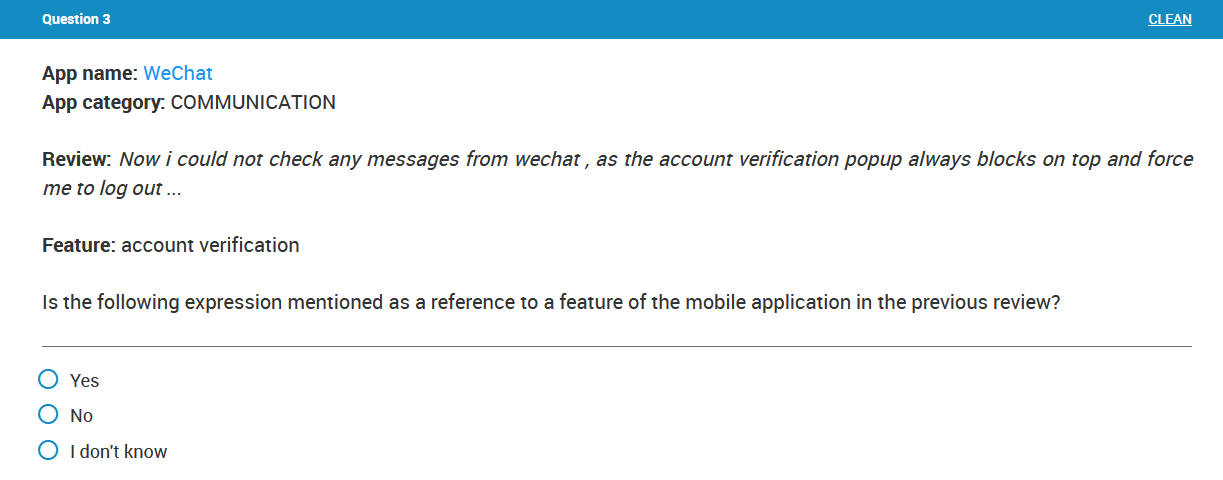}}
    \vskip 6pt
    \caption{Questionnaire for automatic feature extraction ($Q_{A}$)}
    \label{fig:eval-design-fp}
    \end{figure*}
    
    \begin{figure*}[t]
    \centerline{\includegraphics[width=\textwidth]{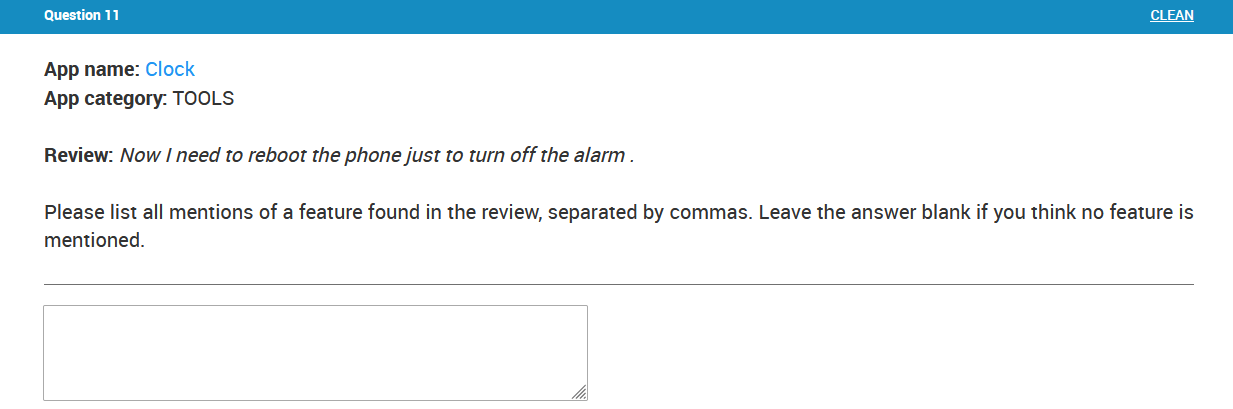}}
    \vskip 6pt
    \caption{Questionnaire for manual feature extraction ($Q_{M}$)}
    \label{fig:eval-design-mfe}
    \end{figure*}

    \item \textbf{Guidelines elaboration.} We prepared annotation instructions for each questionnaire. These include (1) the definition of a feature, (2) the context of the evaluation task, (3) the metadata provided for each annotation task, and (4) several examples with different feature annotations. Guidelines were refined during 3 iterative internal annotations to improve the clarity and representativeness of the examples provided.
    \item \textbf{Evaluation}. We conducted three human evaluation iterations:
    \begin{enumerate}
        \item \textbf{F-measure weighting factor $\beta$ ($Q_A$ and $Q_M$).} 
        We randomly selected a subset of 100 reviews from the evaluation set. For this subset, we created one $Q_A$ questionnaire with 100 reviews and four $Q_M$ questionnaires with 25 reviews each. 
        We measured the time required for each annotator to complete the questionnaire, and we report average values among all valid annotators to measure $A_t$ and $A_T$, which were then used to compute $\beta = A_T / A_t$.\label{human-eval:3} 
        \item \textbf{Precision of new features ($Q_A$ with baseline).} We use the best T-FREX baseline model (RQ\(_1\)) and we run it for inference with all reviews available in the evaluation dataset. We selected only those review-feature annotation pairs where the given feature was not originally annotated in the ground truth. Then, we split the set of filtered review-feature pairs into a subset of questionnaires $Q_A$ according to the pre-defined size (100 reviews).  
        We required a minimum of 5 valid annotators per task. Among these, we used a voting mechanism to determine the final label assigned to each feature. 
        \label{human-eval:1}
        \item \textbf{Precision of new features ($Q_A$ with combined extensions).} We repeated the same process as in~\ref{human-eval:1} but with features predicted by the best-performing T-FREX model with combined extensions (RQ\(_4\)).\label{human-eval:2}
    \end{enumerate}
\end{enumerate}

\subsubsection{Performance efficiency}\label{sec:eval-design-pe-tb}

For the \textit{time behaviour} dimension, we focus on the evaluation of the performance efficiency of the fine-tuning processes for generating T-FREX LLM instances, especially for the assessment of the cost-effectiveness balance with different data partitions (RQ\(_3\)). Consequently, we focus on measuring execution times\footnote{Experiments were conducted on two NVIDIA GeForce RTX 4090 GPUs.
} for each of the fine-tuning stages defined in Section~\ref{sec:baseline-fine-tuning} (i.e., data processing, model loading, training setting, training, and evaluation).
We exclude document pre-processing and feature transfer from performance efficiency analysis as these steps are only executed once before all experiments. 



\subsection{Dataset}\label{sec:dataset}

Table~\ref{tab:dataset} reports the details of the dataset used for evaluation\footnote{Datasets and source code for replicating the evaluation process are available in the GitHub repository: \url{https://github.com/gessi-chatbots/t-frex}}, collected and generated during T-FREX baseline design (see Section~\ref{sec:data-pre-processing}). This includes the mobile app categories included in the dataset, which covers heterogeneous categories ranging from generic \textit{Communication} and \textit{Social} apps to specialized domains such as \textit{Health and fitness} or \textit{Maps and navigation}. Reviews included in this dataset pertain exclusively to those with at least one feature mentioned. 

As a summary, our dataset is composed of 23,816 reviews from 468 mobile apps, leading to 475,382 tokens with the following token class distribution: 29,383 tokens labelled as \textit{B-feature}, 2,841 tokens labelled as \textit{I-feature}, and 443,158 tokens labelled as \textit{O}. The largest category is \textit{Productivity}, with almost 150,000 tokens and up to 77 distinct features ($\mid$features$\mid$).  On the other hand, the least represented categories are \textit{Maps and navigation}, \textit{Lifestyle} and \textit{Weather}, depending on whether we focus on the number of tokens, number of features or distinct features. Notice that a distinct feature might be present in more than one category. For instance, \textit{video calling} is annotated as a feature for both \textit{Productivity} and \textit{Communication} apps.

\begin{table*}[h]
\centering
\caption{Evaluation dataset. List of app categories: \textit{Productivity} (PR), \textit{Communication} (CO), \textit{Tools} (TO), \textit{Social} (SO), \textit{Health and fitness} (HE), \textit{Personalization} (PE), \textit{Travel and local} (TR), \textit{Maps and Navigation} (MA), \textit{Lifestyle} (LI), \textit{Weather} (WE).}
\resizebox{\textwidth}{!}{%
\begin{tabular}{lrrrrrrrrrrr}
\hline
\textbf{metric} &
  \multicolumn{1}{r}{\textbf{PR}} &
  \multicolumn{1}{r}{\textbf{CO}} &
  \multicolumn{1}{r}{\textbf{TO}} &
  \multicolumn{1}{r}{\textbf{SO}} &
  \multicolumn{1}{r}{\textbf{HE}} &
  \multicolumn{1}{r}{\textbf{PE}} &
  \multicolumn{1}{r}{\textbf{TR}} &
  \multicolumn{1}{r}{\textbf{MA}} &
  \multicolumn{1}{r}{\textbf{LI}} &
  \multicolumn{1}{r}{\textbf{WE}} &
  \multicolumn{1}{r}{\textbf{Total}} \\ \hline
\textbf{apps}               & 137     & 51      & 58     & 14     & 75     & 6     & 19     & 31    & 12    & 65     & \textbf{468}     \\
\textbf{reviews}            & 7,348   & 7,003   & 4,321  & 819    & 2,154  & 112   & 530    & 284   & 344   & 901    & \textbf{23,816}  \\
\textbf{sentences}          & 8,604   & 8,135   & 5,402  & 899    & 2,330  & 118   & 602    & 315   & 391   & 984    & \textbf{27,780}  \\
\textbf{tokens}             & 148,172 & 134,833 & 93,395 & 15,597 & 40,907 & 2,022 & 11,105 & 5,868 & 8,044 & 15,439 & \textbf{475,382} \\
\textbf{\textit{B-feature}} & 8,801 & 10,026 & 5,220 & 1,016 & 1,981 & 111 & 691 & 355 & 346 & 836 & \textbf{29,383} \\
\textbf{\textit{I-feature}} & 1,495 & 820 & 305 & 60 & 59 & 1 & 17 & 13 & 47 & 24 & \textbf{2,841} \\
\textbf{\textit{O}} & 137,876 & 123,987 & 87,870 & 14,521 & 38,867 & 1,910 & 10,397 & 5,500 & 7,651 & 14,579 & \textbf{443,158} \\
\textbf{features}           & 8,801 & 10,026 & 5,220 & 1,016 & 1,981 & 111 & 691 & 355 & 346 & 836 & \textbf{29,383} \\ 
\textbf{$\mid$features$\mid$} & 77      & 54      & 50     & 26     & 23     & 19    & 17     & 12    & 10    & 7      & \textbf{198}     \\ 
\hline
\end{tabular}%
}
\label{tab:dataset}
\end{table*}

\subsection{Results}

We structure evaluation results in alignment with research questions, presented as follows: T-FREX baseline fine-tuning (RQ\(_1\)); T-FREX with extended pre-training (RQ\(_2\)); T-FREX with instance selection (RQ\(_3\)); and T-FREX with combined extensions (RQ\(_4\)).

\subsubsection{Baseline fine-tuning}\label{sec:evaluation-baseline}

Table~\ref{tab:token-classification-results} reports average precision ($p$), recall ($r$), standard f-measure ($f_1$) and weighted f-measure ($f_\beta$) for token classification for both out-of-domain\footnote{Based on the scope of this research, we limit the out-of-domain analysis to average results, excluding category-oriented evaluation details. These details are extended in the previous work from which this research stems~\citep{Motger2024saner}.} and in-domain data preparation settings. As mentioned in Section~\ref{sec:evaluation-design}, $\beta$ is computed comparing the performance of $A_T$ (time required for assessing automatic feature extraction) with respect to $A_t$ (time required for manual feature extraction). 
Average results from human evaluation (\ref{human-eval:3}) led to $A_T = 28.29s$ and $A_t = 11.86s$, which leads to a $\beta = A_T / A_t = 2.385$. 
We use $f_\beta$ as the gold metric to select the best-performing models in our research context.

\begin{table*}[h]
\centering
\caption{Token classification results (T-FREX baseline)}
\begin{tabular}{@{}l|rrrr|rrrr@{}}
\toprule
\textbf{} &
  \multicolumn{4}{c}{\textbf{Out-of-domain}} & \multicolumn{4}{c}{\textbf{In-domain}} \\ 
\textbf{model} &
  \textbf{$p$} &
  \textbf{$r$} &
  \textbf{$f_1$} &
  \textbf{$f_\beta$} &
  \textbf{$p$} &
  \textbf{$r$} &
  \textbf{$f_1$} &
  \textbf{$f_\beta$}\\ 
\midrule
\textbf{BERT\textsubscript{base}}       & 0.546 & 0.314 & 0.381 & 0.335 & 0.596 & 0.488 & 0.532 & 0.502 \\
\textbf{BERT\textsubscript{large}}      & 0.577 & 0.339 & 0.414 & 0.361 & 0.719 & \textbf{0.582} & 0.637 & 0.595 \\
\textbf{RoBERta\textsubscript{base}}    & 0.531 & 0.336 & 0.386 & 0.356 & 0.668 & 0.569 & 0.611 & 0.582 \\
\textbf{RoBERTa\textsubscript{large}}   & 0.455 & 0.339 & 0.374 & 0.352 & 0.688 & 0.509 & 0.571 & 0.530 \\
\textbf{XLNet\textsubscript{base}}      & 0.627 & \textbf{0.482} & \textbf{0.535} & \textbf{0.499} & 0.679 & 0.519 & 0.582 & 0.538 \\
\textbf{XLNet\textsubscript{large}}     & \textbf{0.651} & 0.374 & 0.437 & 0.399 & \textbf{0.761} & 0.573 & \textbf{0.646} & \textbf{0.599} \\
\bottomrule
\end{tabular}
\label{tab:token-classification-results}
\end{table*}

For out-of-domain analysis, XLNet\textsubscript{base} outperformed all other models with a recall $r = 0.482$ and a weighted f-measure $f_\beta = 0.499$ ($+0.100$ with respect to the second best-performing model, XLNet\textsubscript{large}). However, the highest precision is reported by XLNet\textsubscript{large} with $p = 0.651$. On the other hand, RoBERTa\textsubscript{large} demonstrated the weakest performance for almost every metric, especially precision. For in-domain analysis, XLNet\textsubscript{large} achieved the highest performance with a precision $p = 0.761$ and weighted f-measure of $f_\beta = 0.599$. BERT\textsubscript{large} reports similar results, especially due to its highest recall with $r = 0.582$ and the weight of recall in computing the weighted f-measure, leading to $f_\beta = 0.595$ (only $-0.005$ with respect to XLNet\textsubscript{large}). 
Conversely, BERT\textsubscript{base} exhibited the lowest performance metrics, with a weighted f-measure $f_\beta = 0.502$.

\begin{table*}[h]
\centering
\caption{Feature extraction evaluation (comparison with SAFE baseline)}
\begin{tabular}{@{}l|rrrr|rrrr@{}}
\toprule
\textbf{} &
  \multicolumn{4}{c}{\textbf{Out-of-domain}} & \multicolumn{4}{c}{\textbf{In-domain}} \\ 
\textbf{model} &
  \textbf{$p$} &
  \textbf{$r$} &
  \textbf{$f_1$} &
  \textbf{$f_\beta$} &
  \textbf{$p$} &
  \textbf{$r$} &
  \textbf{$f_1$} &
  \textbf{$f_\beta$}\\ 
\midrule
\textbf{SAFE}                           & 0.301 & 0.321 & 0.310 & 0.318 & 0.193 & 0.215 & 0.199 & 0.209 \\
\textbf{BERT\textsubscript{base}}       & 0.471 & 0.300 & 0.347 & 0.311 & 0.575 & 0.419 & 0.485 & 0.436 \\
\textbf{XLNet\textsubscript{large}}     & \textbf{0.503} & \textbf{0.417} &\textbf{ 0.445} & \textbf{0.424} & \textbf{0.631} & \textbf{0.572} & \textbf{0.600} & \textbf{0.572} \\
\bottomrule
\end{tabular}
\label{tab:feature-extraction-results-SAFE}
\end{table*}

In addition to token-level effectiveness, we also report and measure effectiveness at the feature level. In this setting, quality metrics are measured using the feature as a whole (i.e., groups of contiguous tokens \textit{B-feature} and \textit{I-feature} composing a whole feature as labelled in the ground truth). This analysis facilitates comparison with SAFE~\citep{Johann2017}, a syntactic-based method in the field of feature extraction which we identify as a baseline method from related work (see Section~\ref{sec:related-work}). Furthermore, this analysis is also intended to facilitate comparisons by further research in the field of feature extraction. We build on a replication of SAFE to support this comparative analysis with our dataset~\citep{Shah2019}. Table~\ref{tab:feature-extraction-results-SAFE} reports functional correctness results for the SAFE approach, BERT\textsubscript{base} (used as the baseline for LLM-based feature extraction design) and XLNet\textsubscript{large} (reported as the best-performing model in T-FREX baseline for token-level effectiveness). Results showcase that T-FREX outperforms SAFE in all settings, with the only exception of the out-of-domain recall with BERT\textsubscript{base}. For in-domain analysis, T-FREX reports significantly greater precision ($+0.438$) and recall ($+0.357$) with respect to syntactic-based mechanisms, especially for the former. Overall, in-domain T-FREX baseline version showcases to overcome some of the limitations posed by syntactic-based approaches in the context of reviews, especially when generalizing syntactic patterns to different datasets, as suggested by~\cite{Shah2019}.

Table~\ref{tab:new-features-baseline} summarizes the results of the human evaluation of new features (\ref{human-eval:1}). We collected all features predicted by XLNet\textsubscript{large} fine-tuned model during the in-domain $k$-fold cross-validation analysis for each test set. Then, we selected those reviews with predicted features which were not originally annotated as features in the ground-truth set. This led to a total of 1,956 reviews, with 1,067 distinct feature annotations. Given the size of the dataset, we decided to submit for evaluation all reviews, leading to 21 human evaluation tasks of 100 feature annotation questions (95 for evaluation, 5 for control). As a result, human evaluation of new features leads to a total average precision of 0.625 (i.e., 62.5\% of `\textit{Yes}' annotations across the whole dataset). This supports the hypothesis that the original dataset lacks exhaustive annotations.

\begin{table}[h]
\caption{Human evaluation of new features (FP) with best-performing T-FREX baseline model (XLNet\textsubscript{large}). 
}
\label{tab:new-features-baseline}
\resizebox{\textwidth}{!}{%
\begin{tabular}{@{}lrrrrrrrrrrr@{}}
\toprule
\textbf{} & \textbf{PR} & \textbf{CO} & \textbf{TO} & \textbf{SO} & \textbf{HE} & \textbf{PE} & \textbf{TR} & \textbf{MA} & \textbf{LI} & \textbf{WE} & \textbf{Total} \\ \midrule
\textbf{\#reviews} & 459    & 643    & 560    & 44     & 218    & 0 & 8      & 29     & 0 & 0 & \textbf{1,956}   \\
\textbf{\% Yes}    & 68.6\% & 62.3\% & 58.4\% & 63.6\% & 59.4\% & - & 66.7\% & 58.6\% & - & - & \textbf{62.5\%} \\
\textbf{\% No}     & 28.8\% & 35.0\% & 41.7\% & 34.1\% & 39.3\% & - & 33.3\% & 41.4\% & - & - & \textbf{36.1\%} \\
\textbf{\% Idk}    & 1.6\%  & 2.7\%  & 1.8\%  & 2.2\%  & 0.6\%  & - & 0.0\%  & 0.0\%  & - & - & \textbf{1.9\%}  \\
\bottomrule
\end{tabular}%
}
\end{table}

\subsubsection{Extended pre-training}\label{sec:eval-extended-pretraining}

\begin{figure}[h]
    \centering
    \includegraphics[width=0.8\textwidth]{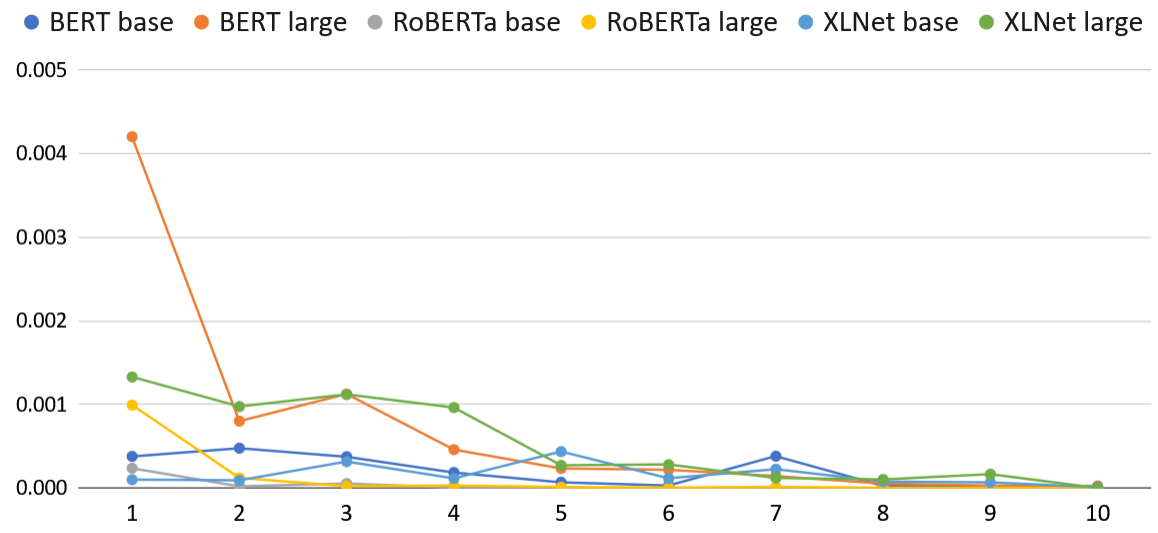} 
    \caption{Evaluation loss across 10 epochs.}
    \label{fig:evaluation-loss}
\end{figure}

Figure~\ref{fig:evaluation-loss} showcases the evolution of the evaluation loss during the extended pre-training stage after each epoch $1 \rightarrow 10$. All models, especially large model instances (i.e., BERT\textsubscript{large}, RoBERTa\textsubscript{large}, XLNet\textsubscript{large}) show a general trend of decreasing evaluation loss over the epochs, with the largest reduction occurring between the first and second epochs. On the other hand, base models (i.e., BERT\textsubscript{base}, RoBERTa\textsubscript{base}, XLNet\textsubscript{base}) exhibit relatively stable and lower evaluation losses throughout the epochs. Between epochs $8 \rightarrow 10$ all six models converge into evaluation loss values $\leq 10^{-4}$. The higher initial evaluation loss for large models may be attributed to their increased complexity and greater number of parameters, which require more epochs for effective optimization.

\begin{figure}[t]
    \centering
    \begin{subfigure}[b]{0.49\textwidth}
        \centering
        \includegraphics[width=\textwidth]{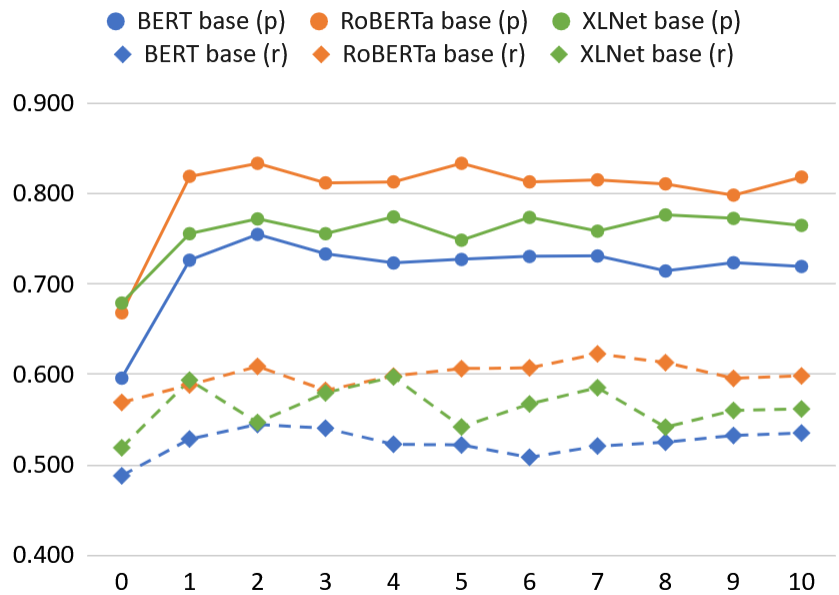}
        \caption{Base models}
        \label{fig:results-EP-base}
    \end{subfigure}
    \hfill
    \begin{subfigure}[b]{0.49\textwidth}
        \centering
        \includegraphics[width=\textwidth]{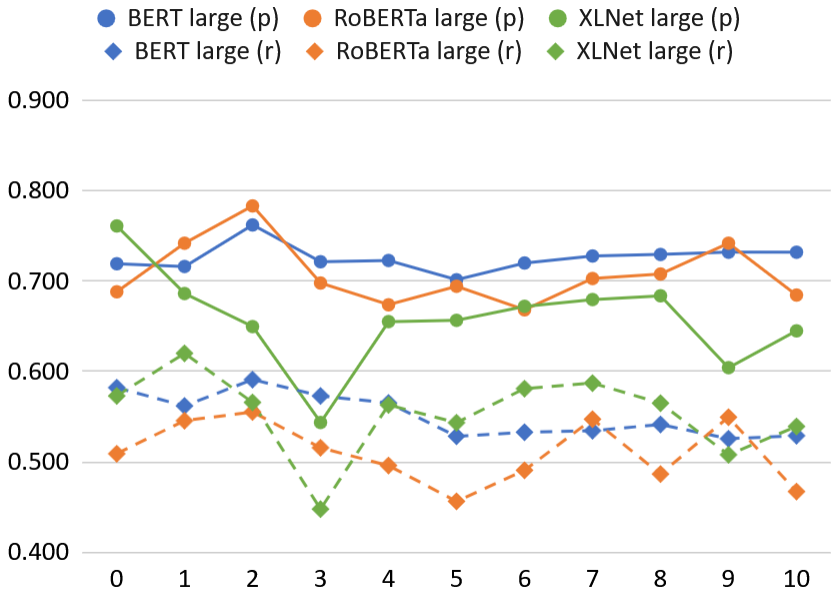}
        \caption{Large models}
        \label{fig:results-EP-large}
    \end{subfigure}
    \caption{Precision ($p$) and recall ($r$) evolution with extended pre-training.}
    \label{fig:results-EP}
\end{figure}

Figure~\ref{fig:results-EP} illustrates the evolution of precision and recall metrics for the in-domain fine-tuning using model checkpoints $c$ from $1 \rightarrow 10$. For a comparative analysis, we include T-FREX baseline model as model checkpoint $c = 0$. Results showcase that all six model instances increase the maximum value for each metric at some point during the extended pre-training. Precision is on average the most increased metric, especially for base models BERT\textsubscript{base} ($c=2$, +0.159), RoBERTa\textsubscript{base} ($c=5$, +0.166) and XLNet\textsubscript{base} ($c=2$, +0.097). The only exception is precision for XLNet\textsubscript{large}, which suffers from decay from 0.761 to 0.686 (-0.075). The increase of recall is more modest and is especially highlighted in base models like XLNet\textsubscript{base} ($c=4$, +0.078) but also in large models such as RoBERTa\textsubscript{large} ($c=9$, +0.040) or XLNet\textsubscript{large} ($c=1$, +0.047). 

If we focus on evolution across epochs, we notice that most models approach the best metric values between the first and the second epoch. After that, metric values either stabilize (i.e., BERT\textsubscript{base}, RoBERTa\textsubscript{base}, XLNet\textsubscript{base}) or start to decay (i.e., BERT\textsubscript{base}, RoBERTa\textsubscript{large}). In addition, precision and recall values show a common behaviour across all epochs (i.e., increasing one also increases the other). The only exception to both of these statements is XLNet\textsubscript{large}. Using XLNet\textsubscript{large} extended with just one epoch increases recall (+0.047) and, consequently, $f_\beta$ (+0.036). But given the decay in precision, the balanced f-measure remains almost identical (+0.002). Finally, in terms of evolutionary behaviour, BERT (both base and large) and RoBERTa\textsubscript{base} show relatively stable behaviour between consecutive epochs. However, RoBERTa\textsubscript{large} and XLNet (both base and large) showcase erratic behaviour with constants increases and decays. This behaviour might be a consequence of the pre-training objectives and the data used for the initial pre-training of these models (see Section~\ref{sec:discussion}).

\subsubsection{Instance selection}\label{sec:eval-instance-selection}

\begin{figure}[t]
    \centering
    \begin{subfigure}[b]{0.49\textwidth}
        \centering
        \includegraphics[width=\textwidth]{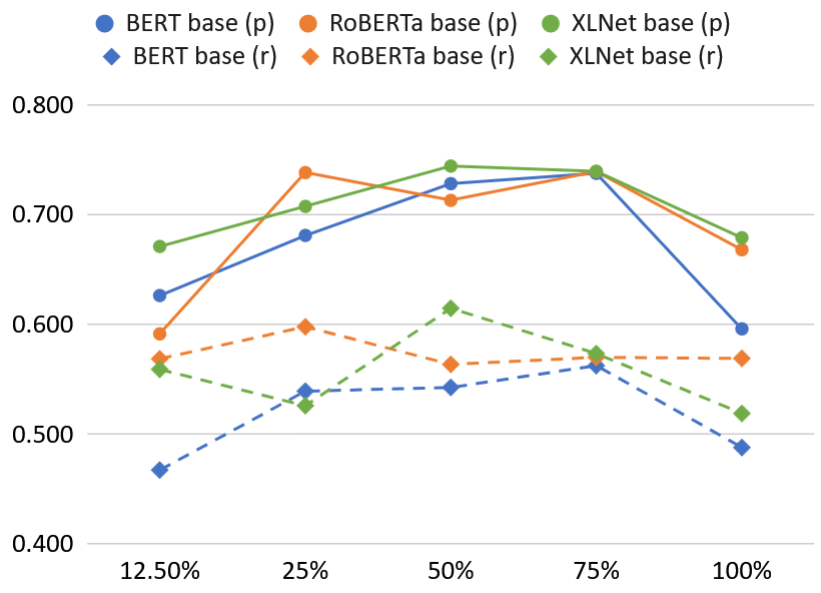}
        \caption{Base models}
        \label{fig:results-IS-base}
    \end{subfigure}
    \hfill
    \begin{subfigure}[b]{0.49\textwidth}
        \centering
        \includegraphics[width=\textwidth]{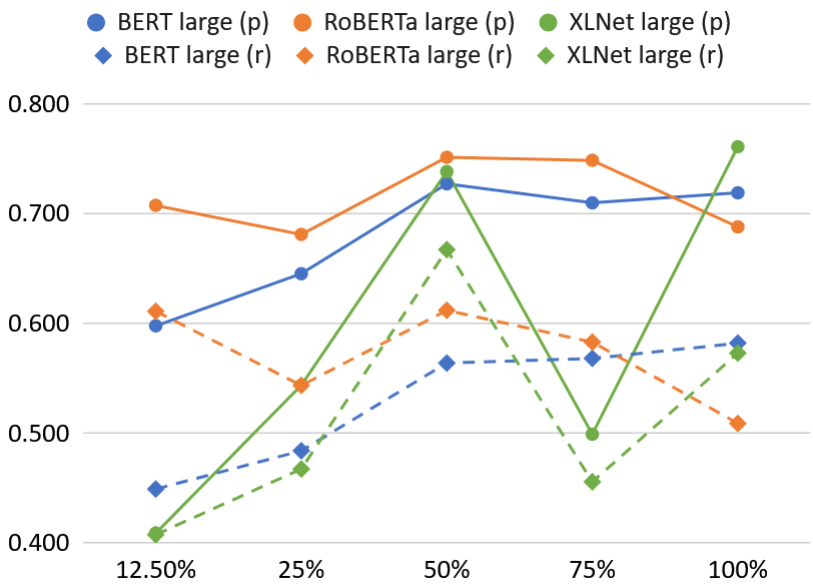}
        \caption{Results for BERT\textsubscript{large}}
        \label{fig:results-IS-large}
    \end{subfigure}
    
    \caption{Precision ($p$) and recall ($r$) evolution with instance selection.}
    \label{fig:results-IS}
\end{figure}

Figure~\ref{fig:results-IS} illustrates the evolution of precision and recall metrics for the in-domain fine-tuning using T-FREX baseline approach for fine-tuning and the instance selection algorithm to generate different training data set partitions $d \in \{12.5\%, 25\%, 50\%, 75\%\}$. For a comparative, evolutionary analysis, we include T-FREX baseline setting using the complete training set as $d = 100\%$.

All T-FREX base models experience an improvement in every metric when a certain degree of instance selection (i.e., between 12.5\% and 75\%) is conducted in the ground truth training dataset. For BERT\textsubscript{base}, the best data partition is 75\% ($f_\beta = 0.583$). For RoBERTa\textsubscript{base}, using only 25\% leads to the best results ($f_\beta = 0.615$). For XLNet\textsubscript{base}, the best data partition is 50\% ($f_\beta = 0.631$). While this condition mostly prevails in large models, BERT\textsubscript{large} (for recall and f-measures) and XLNet\textsubscript{large} (for precision) report some particular exceptions. Beyond these, using 50\% of the training data results in the best setting for RoBERTa\textsubscript{large} ($f_\beta = 0.629$) and XLNet\textsubscript{large} ($f_\beta = 0.677$). On average, for six models and two fundamental metrics (precision and recall), 10 out of 12 evaluations improve with instance selection. Specifically, two model-metric combinations improve with 25\% of the data (RoBERTa\textsubscript{base}, precision and recall), two combinations improve with 75\% (BERT\textsubscript{base}, precision and recall), and the rest improve with a 50\% partition.

If we focus on the tendency as we increase the size of the training set, we observe a non-linear behaviour that suggests the presence of a local minimum. This indicates an optimal value for the training set size where the model performance is maximized before it starts to decline with further data increase. This phenomenon is consistent across various models and metrics. For instance, for BERT\textsubscript{base} and RoBERTa\textsubscript{base}, the metrics peak at different points — 75\% and 25\% respectively — before showing a decline, which highlights the importance of instance selection in optimizing model performance. Similarly, for large models like BERT\textsubscript{large} and XLNet\textsubscript{large}, the optimal training set sizes are different, with 50\% being optimal for RoBERTa\textsubscript{large} and XLNet\textsubscript{large}.

\begin{figure*}[h]
\centerline{\includegraphics[width=\textwidth]{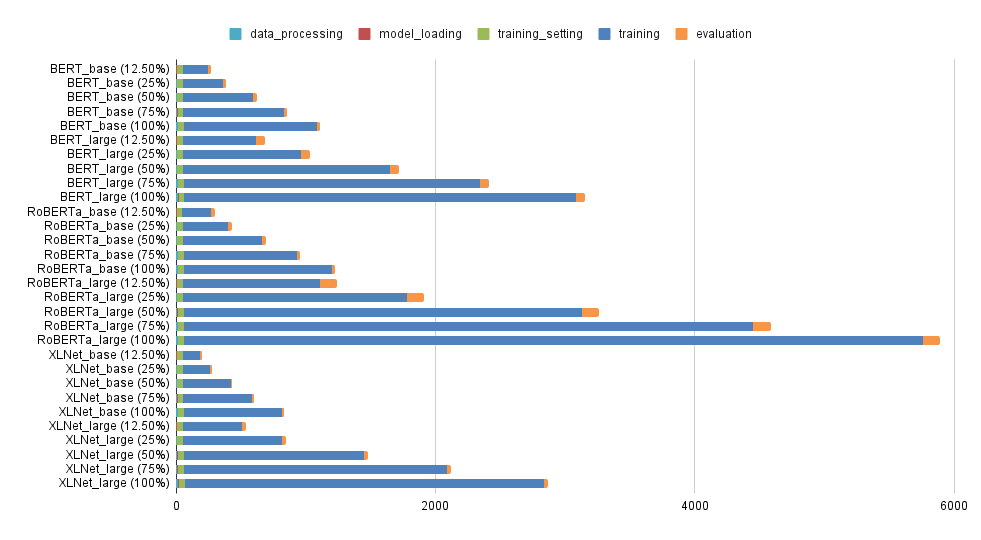}}
\vskip 6pt
\caption{Execution times (in seconds) for T-FREX model fine-tuning.}
\label{fig:execution-times}
\end{figure*}

Concerning variations on time behaviour using different data partitions, Figure~\ref{fig:execution-times} reports, for each model and training data partition, the execution times required for each model fine-tuning stage (steps $ 1 \rightarrow 6$ in Section~\ref{sec:model-fine-tuning}). We report average values obtained from the $k$-fold cross-validation during the in-domain analysis. The training stage takes the majority of the total execution time across all models and data partitions. In comparison, data processing and model loading phases consume relatively minimal time. On average, using 50\% of the training set entails a speed up of $\times1.8$ with respect to using 100\% of the dataset. For smaller partitions, the speed up grows linearly. On average, for base models, using 12.5\% of the training set entails a $\times4.1$ speed up, while for large models this is increased to $\times4.9$ on average. If we focus on larger data partitions, for 75\%, all models consistently report a speedup of $\times1.3$.

\subsubsection{Combining extended pre-training and instance selection}\label{sec:eval-combined}

\begin{table}[t]
\caption{Metrics for the best-performing approach with background colors representing $\delta$ variations relative to the T-FREX baseline (BL). Bolded cells indicate the best metrics for each model. Pinpointed cells \bestmark{} denote the overall best metrics across all evaluations.}
\resizebox{\textwidth}{!}{%
\begin{tabular}{@{}ll>{\columncolor[HTML]{FFFFFF}}r>{\columncolor[HTML]{FFFFFF}}r>{\columncolor[HTML]{FFFFFF}}r>{\columncolor[HTML]{FFFFFF}}r@{}}
\toprule
\textbf{model} & \textbf{setting} & \multicolumn{1}{r}{\textbf{$p$}} & \multicolumn{1}{r}{\textbf{$r$}} & \multicolumn{1}{r}{\textbf{$f_1$}} & \multicolumn{1}{r}{\textbf{$f\textsubscript{$\beta$}$}} \\
\midrule
\multirow{4}{*}{\textbf{BERT\textsubscript{base}}}    
& \textbf{BL}    & 0.596             & 0.488             & 0.532             & 0.502             \\
& \textbf{EP}    & \cellcolor{green!92}\textsuperscript{c=2}0.755 & \cellcolor{green!33}\textsuperscript{c=2}0.545 & \cellcolor{green!52}\textsuperscript{c=2}0.621 & \cellcolor{green!39}\textsuperscript{c=2}0.569 \\
& \textbf{IS}    & \cellcolor{green!83}\textsuperscript{d=75\%}0.738 & \cellcolor{green!43}\textbf{\textsuperscript{d=75\%}0.562} & \cellcolor{green!58}\textbf{\textsuperscript{d=75\%}0.632} & \cellcolor{green!47}\textbf{\textsuperscript{d=75\%}0.583} \\
& \textbf{EP/IS} & \cellcolor{green!100}\textbf{\textsuperscript{c=2,d=75\%}0.768} & \cellcolor{green!35}\textsuperscript{c=1,d=75\%}0.549 & \cellcolor{green!55}\textsuperscript{c=1,d=75\%}0.626 & \cellcolor{green!41}\textsuperscript{c=1,d=75\%}0.573 \\
\midrule

\multirow{4}{*}{\textbf{BERT\textsubscript{large}}}    
& \textbf{BL}    & 0.719             & 0.582             & 0.637             & 0.599             \\
& \textbf{EP}    & \cellcolor{green!25}\textsuperscript{c=2}\textbf{0.762} & \cellcolor{green!29}\textsuperscript{c=2}\textbf{0.632} & \cellcolor{green!31}\textsuperscript{c=2}\textbf{0.691} & \cellcolor{green!29}\textsuperscript{c=2}\textbf{0.649} \\
& \textbf{IS}    & \cellcolor{green!5}\textsuperscript{d=50\%}0.727 & \cellcolor{red!8}\textsuperscript{d=75\%}0.568 & \cellcolor{red!6}\textsuperscript{d=75\%}0.626 & \cellcolor{red!7}\textsuperscript{d=75\%}0.587 \\
& \textbf{EP/IS} & \cellcolor{red!15}\textsuperscript{c=7,d=75\%}0.692 & \cellcolor{green!2}\textsuperscript{c=4,d=75\%}0.587 & \cellcolor{red!4}\textsuperscript{c=4,d=75\%}0.630 & \cellcolor{green!1}\textsuperscript{c=4,d=75\%}0.601 \\
\midrule

\multirow{4}{*}{\textbf{RoBERTa\textsubscript{base}}}  
& \textbf{BL}    & 0.668             & 0.569             & 0.611             & 0.582             \\
& \textbf{EP}    & \cellcolor{green!97}\bestmark{\textsuperscript{c=5}0.834} & \cellcolor{green!31}\textsuperscript{c=7}\textbf{0.623} & \cellcolor{green!49}\textsuperscript{c=7}\textbf{0.694} & \cellcolor{green!38}\textsuperscript{c=7}\textbf{0.647} \\
& \textbf{IS}    & \cellcolor{green!41}\textsuperscript{d=75\%}0.739 & \cellcolor{green!17}\textsuperscript{d=25\%}0.598 & \cellcolor{green!23}\textsuperscript{d=25\%}0.652 & \cellcolor{green!19}\textsuperscript{d=25\%}0.616 \\
& \textbf{EP/IS} & \cellcolor{green!42}\textsuperscript{c=5,d=75\%}0.741 & \cellcolor{green!13}\textsuperscript{c=5,d=25\%}0.592 & \cellcolor{green!22}\textsuperscript{c=7,d=75\%}0.650 & \cellcolor{green!16}\textsuperscript{c=7,d=75\%}0.610 \\
\midrule

\multirow{4}{*}{\textbf{RoBERTa\textsubscript{large}}} 
& \textbf{BL}    & 0.688             & 0.509             & 0.571             & 0.530             \\
& \textbf{EP}    & \cellcolor{green!55}\textsuperscript{c=2}\textbf{0.783} & \cellcolor{green!26}\textsuperscript{c=2}0.555 & \cellcolor{green!37}\textsuperscript{c=2}0.636 & \cellcolor{green!29}\textsuperscript{c=2}0.580 \\
& \textbf{IS}    & \cellcolor{green!36}\textsuperscript{d=50\%}0.751 & \cellcolor{green!59}\textsuperscript{d=50\%}0.612 & \cellcolor{green!55}\textsuperscript{d=50\%}\textbf{0.666} & \cellcolor{green!57}\textsuperscript{d=50\%}0.629 \\
& \textbf{EP/IS} & \cellcolor{green!47}\textsuperscript{c=9,d=75\%}0.769 & \cellcolor{green!65}\textsuperscript{c=1,d=12.5\%}\textbf{0.621} & \cellcolor{green!54}\textsuperscript{c=1,d=12.5\%}0.664 & \cellcolor{green!63}\textsuperscript{c=1,d=12.5\%}\textbf{0.639} \\
\midrule

\multirow{4}{*}{\textbf{XLNet\textsubscript{base}}}    
& \textbf{BL}    & 0.679             & 0.519             & 0.582             & 0.538             \\
& \textbf{EP}    & \cellcolor{green!56}\textsuperscript{c=8}0.776 & \cellcolor{green!45}\textsuperscript{c=4}0.597 & \cellcolor{green!45}\textsuperscript{c=4}0.661 & \cellcolor{green!46}\textsuperscript{c=4}0.618 \\
& \textbf{IS}    & \cellcolor{green!37}\textsuperscript{d=50\%}0.744 & \cellcolor{green!55}\textsuperscript{d=50\%}\textbf{0.615} & \cellcolor{green!47}\textsuperscript{d=50\%}\textbf{0.664} & \cellcolor{green!54}\textsuperscript{d=50\%}\textbf{0.631} \\
& \textbf{EP/IS} & \cellcolor{green!60}\textsuperscript{c=2,d=75\%}\textbf{0.783} & \cellcolor{green!44}\textsuperscript{c=3,d=75\%}0.595 & \cellcolor{green!41}\textsuperscript{c=2,d=75\%}0.655 & \cellcolor{green!40}\textsuperscript{c=2,d=75\%}0.617 \\
\midrule

\multirow{4}{*}{\textbf{XLNet\textsubscript{large}}}   
& \textbf{BL}    & \textbf{0.761}             & 0.573             & 0.646             & 0.595             \\
& \textbf{EP}    & \cellcolor{red!13}\textsuperscript{c=1}0.738 & \cellcolor{green!55}\textsuperscript{c=1}\bestmark{0.667} & \cellcolor{green!31}\textsuperscript{c=1}\bestmark{0.700} & \cellcolor{green!48}\textsuperscript{c=1}\bestmark{0.677} \\
& \textbf{IS}    & \cellcolor{red!44}\textsuperscript{d=50\%}0.686 & \cellcolor{green!27}\textsuperscript{d=50\%}0.620 & \cellcolor{green!1}\textsuperscript{d=50\%}0.648 & \cellcolor{green!20}\textsuperscript{d=50\%}0.629 \\
& \textbf{EP/IS} & \cellcolor{red!8}\textsuperscript{c=10,d=75\%}0.748 & \cellcolor{green!40}\textsuperscript{c=7,d=25\%}0.641 & \cellcolor{green!13}\textsuperscript{c=7,d=25\%}0.669 & \cellcolor{green!35}\textsuperscript{c=7,d=25\%}0.655 \\
\bottomrule
\end{tabular}%
}
\label{tab:best-settings}
\end{table}

Given the large number of experimentation settings for combined analysis (6 T-FREX models $\times$ 4 data partitions $\times$ 11 model checkpoints), we limit the results included in this paper to the best configuration of extended pre-training and/or instance selection for each metric. To this end, Table~\ref{tab:best-settings} reports functional correctness metrics for the token classification task with respect to T-FREX baseline (RQ\(_1\)), T-FREX with extended pre-training (RQ\(_2\)), T-FREX with instance selection (RQ\(_3\)) and T-FREX with combined extensions (RQ\(_4\)). For each combination, we exclusively report the setting reporting the best metric by specifying the model checkpoint ($c$) for extended pre-training, the data partition ($d$) for instance selection, and both when combining extensions.

If we focus on the combined use of instance selection and extended pre-training (EP/IS), there are only a few examples where this combination is the best option for functional correctness: precision for BERT\textsubscript{base} and XLNet\textsubscript{base}; and recall and $f_\beta$ for RoBERTa\textsubscript{large}. If we focus on the use of extended pre-training only (EP), BERT\textsubscript{large}, RoBERTa\textsubscript{base} and XLNet\textsubscript{large} report this design as the best option for increasing effectiveness. In fact, XLNet\textsubscript{large} emerges as the best model, both for recall ($r = 0.700$) and weighted f-measure ($f_\beta = 0.677$), with $c=1$. For precision, RoBERTa\textsubscript{base} with extended pre-training with $c=5$ emerges as the best option. On the other hand, if we focus on the isolated use of instance selection, it emerges as the best option for BERT\textsubscript{base} and RoBERTa\textsubscript{base}, with some exceptions like precision for BERT\textsubscript{base}.

Focusing on delta variations, we observe that on average all base models, as well as mostly all large models, improve T-FREX baseline when extended pre-training and/or instance selection is applied. For large models, there are some minor exceptions like precision for XLNet\textsubscript{large} and f-measures for BERT\textsubscript{large}. Precision is the most increased metric, experiencing its greatest improvement when extended pre-training and instance selection are used in combination for BERT\textsubscript{base} (+0.172). Recall is also increased, but its maximum improvement is more conservative, as observed with RoBERTa\textsubscript{large} with combined extensions (+0.112) or in XLNet\textsubscript{large} with instance selection (+0.094).

After analysis of all T-FREX settings, XLNet\textsubscript{large} with extended pre-training ($c=1$) emerges as the best-performing\footnote{Notice the criteria for ``best-performing model" might vary across different research contexts and scenarios, for which we provide exhaustive results for precision and recall.} model (based on $f_\beta$). For consistency with the evaluation of T-FREX baseline, we complement the analysis of ground truth annotations with the evaluation of new features reported by T-FREX which are not present in the ground truth. We use the fine-tuned \textsuperscript{c=1}XLNet\textsubscript{large} model for inference, collecting 11,120 reviews mentioning 1,311 distinct new features not assessed during evaluation of RQ\(_1\). Given the size of the dataset of reviews, we limited the set of reviews used for human evaluation for resource optimization to 1,311 reviews (i.e., one review instance of each distinct feature). This entails 11.8\% of the complete set of reviews and 100\% of the newly predicted features. Consequently, human evaluation (\ref{human-eval:2}) consisted of 14 tasks with 100 questions (95 for evaluation, 5 for control). 

Table~\ref{tab:new-features-extended} reports the results for this analysis using the same format as in RQ\(_1\) (Table~\ref{tab:new-features-baseline}). On average, we observe a consistent precision of features labelled as `\textit{Yes}' (60.8\%), slightly lower with respect to T-FREX baseline (62.5\%). However, the set of features labelled as `\textit{No}' is -3\% smaller for the extended version (33.1\%) with respect to T-FREX baseline (36.1\%). This is due to a major uncertainty in label predictions annotated during this evaluation process, labelled as `\textit{I don't know}' (6.1\%). In addition, the number of new features reported with extended pre-training (11,120) is $\times 5.7$ higher than the number of new features reported with the baseline design (1,956). Consequently, although precision is not increased, T-FREX with \textsuperscript{c=1}XLNet\textsubscript{large} significantly reduces the number of potentially missed features compared to T-FREX baseline.


\begin{table}[t]
\caption{Human evaluation of new features (FP) with best-performing T-FREX extended model (\textsuperscript{c=1}XLNet\textsubscript{large}). 
}
\label{tab:new-features-extended}
\resizebox{\textwidth}{!}{%
\begin{tabular}{@{}lrrrrrrrrrrr@{}}
\toprule
\textbf{} & \textbf{PR} & \textbf{CO} & \textbf{TO} & \textbf{SO} & \textbf{HE} & \textbf{PE} & \textbf{TR} & \textbf{MA} & \textbf{LI} & \textbf{WE} & \textbf{Total} \\ \midrule
\textbf{\#reviews} & 206    & 228    & 260    & 102    & 282    & 8      & 93     & 46     & 35     & 60     & \textbf{1320}   \\
\textbf{\% Yes}    & 57.8\% & 68.0\% & 60.0\% & 71.6\% & 53.2\% & 87.5\% & 51.6\% & 54.3\% & 88.6\% & 63.3\% & \textbf{60.8\%} \\
\textbf{\% No}     & 35.4\% & 25.9\% & 34.6\% & 22.5\% & 38.7\% & 12.5\% & 43.0\% & 41.3\% & 8.6\%  & 33.3\% & \textbf{33.1\%} \\
\textbf{\% Idk}    & 6.8\%  & 6.1\%  & 5.4\%  & 5.9\%  & 8.2\%  & 0.0\%  & 5.4\%  & 4.3\%  & 2.9\%  & 3.4\%  & \textbf{6.1\%}  \\
\bottomrule
\end{tabular}%
}
\end{table}


\section{Discussion}\label{sec:discussion}

\subsection{Baseline fine-tuning}

\begin{tcolorbox}[colback=teal!5!white, colframe=teal!75!black,left=1mm, right=1mm, top=1mm, bottom=1mm
]
\textbf{RQ\(_1\):} The T-FREX baseline design improves upon syntactic-based baseline methods for extracting feature mentions from mobile app reviews (H$_1$). These improvements are evident in both out-of-domain and in-domain learning settings, with notable gains in precision and recall.
\end{tcolorbox}

\textbf{For the most challenging scenario (i.e., out-of-domain), XLNet\textsubscript{base} demonstrated the strongest performance}, achieving a weighted f-measure ($f_\beta$) of 0.499, aligning with H$_1$ even in less familiar contexts. This model is particularly effective in reducing missed features (false negatives). However, XLNet\textsubscript{large} excelled in precision, achieving the highest score of 0.651, indicating its ability to minimize false features (false positives). Conversely, RoBERTa\textsubscript{large} showed the weakest performance, highlighting its limited generalization capability for out-of-domain scenarios. 

\textbf{For the most common scenario (i.e., in-domain), XLNet\textsubscript{large} outperformed XLNet\textsubscript{base} across all metrics}, with a precision of 0.761 and a weighted f-measure of 0.599, overcoming its out-of-domain limitations. BERT\textsubscript{large} also performed well, especially in recall with a score of 0.582, but slightly lagged in weighted f-measure at 0.595. Despite the small differences, XLNet\textsubscript{large} emerged as the best overall model for in-domain tasks. Notably, larger RoBERTa models performed worse than their base versions, consistent with other studies suggesting that larger models may sometimes hinder performance. 

Human evaluation of new features using XLNet\textsubscript{large} for in-domain analysis revealed an average precision of 0.625, confirming the model's ability to identify valid, previously unannotated features. This supports the hypothesis that the original dataset's annotations were not exhaustive. 

Overall, the \textbf{results highlight the improved effectiveness of T-FREX models, particularly the XLNet variants, in various evaluation contexts}, reinforcing H$_1$ as they outperform traditional syntactic-based methods. An encoder-only LLM approach showcases better adaptation to the concept of a feature as used in an industrial setting like AlternativeTo, suppressing the limitations of deterministic approaches conditioned by the use of specific syntactic patterns that might not always generalize. Hence, we argue that T-FREX provides an effective software-based mechanism to reduce missing features in automatic review-based opinion mining pipelines used by practitioners and researchers.

\subsection{Extended pre-training}

\begin{tcolorbox}[colback=teal!5!white, colframe=teal!75!black, left=1mm, right=1mm, top=1mm, bottom=1mm
]
\textbf{RQ\(_2\):} Extended pre-training consistently improves the precision and recall of feature extraction for all encoder-only models analysed in this research (H$_2$). Different models entail different outcomes, requiring different training epochs according to the size and pre-training task of the model to reach maximum improvement without forcing the decay of quality metrics.
\end{tcolorbox}

\textbf{Base models 
demonstrate substantial improvements in performance metrics with extended pre-training}, with precision showing the most significant increases. This suggests that base models have greater potential for refinement, and extended pre-training effectively reduces false positives, enhancing the likelihood that identified features are accurate. Similarly, while large models tend to also improve with extended pre-training, they also tend to experience a decay after the first few epochs, indicating a limit to the benefits of extended pre-training for these more complex models. This is also supported by the evolution of the evaluation loss, which is significantly limited after a few epochs of extended pre-training.

In addition, \textbf{base models tend to stabilize across epochs in comparison to the erratic behaviour of large models}, especially RoBERTa\textsubscript{large} and XLNet\textsubscript{large}. This instability may be attributed to the complexity of their pre-training tasks and the data used, which could lead to fluctuations in performance as the models are fine-tuned. Specifically, XLNet\textsubscript{large} showcases significant variability, potentially due to its architectural differences and the pre-training objective (i.e., PLM), which might not align as well with the fine-tuning tasks. This erratic behaviour underscores the need for careful monitoring and potentially different strategies when employing large models for extended pre-training.

On average, the \textbf{best performance improvements are achieved after one or two epochs of extended pre-training}, further validating H$_2$ by showing that even limited extended pre-training enhances LLMs' reflection of mobile app review contexts. Beyond this point, we observe similar or even diminishing results. In similar contexts, this suggests that investing in extensive pre-training beyond a few epochs may not be adequate in terms of cost-effectiveness balance, given the diminishing returns and significant computational resources required. Researchers and practitioners conducting similar experiments (e.g., NER fine-tuning with encoder-only LLMs) should consider these factors when designing their training protocols to ensure efficient and effective use of computational resources.

\subsection{Instance selection}

\begin{tcolorbox}[colback=teal!5!white, colframe=teal!75!black, left=1mm, right=1mm, top=1mm, bottom=1mm
]
\textbf{RQ\(_3\):} On average, instance selection not only improves the performance efficiency of the fine-tuning process but also leads to an improvement in the functional correctness of the token classification task, both in precision and recall (H$_3$). This improvement is especially observed for data partitions between 50\%-75\%, and it is consistently observed in base models, while large models present some exceptions.
\end{tcolorbox}

As illustrated with the analysis of base models, \textbf{optimal performance is achieved with different data partitions from the original set of reviews} (e.g., 75\% for BERT\textsubscript{base}, 25\% for Ro\-BERTa\textsubscript{base}, and 50\% for XLNet\textsubscript{base}). This underscores that base models may struggle with redundancies and noise in the full dataset, and a more targeted data selection can enhance their performance. Large models, while also benefiting from instance selection, show some variability, particularly with XLNet\textsubscript{large}, which exhibits erratic behaviour, suggesting that its architecture might be more sensitive to the training set size and content. Interestingly, the performance across all models tends to peak at certain data partition sizes and then decline, indicating a local optimum. Results showcase the validity of the instance selection method proposed for 
token-level fine-tuning in the context of mobile app review feature extraction.

In addition to effectiveness, \textbf{using a smaller, optimal portion of the dataset significantly reduces execution times}. Training with 50\% of the dataset nearly halves the execution time while still improving performance on average. For even smaller partitions, such as 12.5\%, the speedup can be as much as four to five times faster, which is particularly advantageous for large-scale applications where computational resources and time are critical. This efficiency gain, combined with improved model performance, makes instance selection a highly valuable strategy in real-world scenarios, where processing large batches of reviews for multiple tasks (e.g., sentiment analysis, content classification, feature extraction) becomes computationally expensive.


\subsection{Combining extended pre-training and instance selection}

\begin{tcolorbox}[colback=teal!5!white, colframe=teal!75!black, left=1mm, right=1mm, top=1mm, bottom=1mm
]
\textbf{RQ\(_4\):} Combined use of extended pre-training and instance selection improves T-FREX baseline design in almost all design scenarios and models. This entails that a cost-effective assessment is necessary to improve both the effectiveness and efficiency of the system. However, optimal precision and recall values are overall achieved with extended pre-training without the need for instance selection. Results report some exceptions to this, especially for base models, where either combined extensions or simply instance selection becomes the best option.
\end{tcolorbox}


\textbf{When focusing on precision, RoBERTa\textsubscript{base} stands out as the best model}, especially for tasks where false positives are more tolerable than missing actual features. Conversely, \textbf{for applications where recall is critical, XLNet\textsubscript{large} is superior}, as missing a feature mention is more costly than incorrectly identifying one. This underscores the need for decision-making based on specific application scenarios, which is why detailed reporting of these metrics is essential. However, in the context of RE \textit{hairy} tasks~\citep{Berry2021}, we argue that high recall is more important than high precision, which is reflected in the weighting of $f_\beta$. 

On average, \textbf{base models benefit more from instance selection, particularly in terms of recall}, demonstrating their sensitivity to data quality and relevance. On the other hand, large models show significant improvements with extended pre-training, whether used alone or in conjunction with instance selection. Notably, base models always experience improvements with some form of extension. This motivates the need for a cost-effectiveness analysis considering both absolute performance and relative improvements (Table~\ref{tab:best-settings}) to determine the practicality of maintaining base versus large models.

Regarding new, unseen features, \textbf{human evaluation indicates a substantial increase in the number of identified features with extended T-FREX design}. Due to the large number of feature candidates in the dataset (i.e., tokens), a slight improvement in recall ($+0.047$, \textsuperscript{c=1}XLNet\textsubscript{large}) leads to a significant increase in potential new features in absolute numbers (1,956 vs. 11,120). Consequently, despite a slight drop in the precision of new features ($-0.017$), the much higher number of new feature mentions identified suggests a lower likelihood of missing important features. This trade-off supports a more exhaustive feature extraction process, as previously argued.

\subsection{Threats to validity}\label{sec:ttv}

For the elicitation of threats to validity, we rely on the taxonomy proposed by~\cite{Wohlin2014}. Concerning \textit{construct validity}, we rely on functional correctness metrics based on token classification performance at the token level. However, the feature extraction process is not evaluated in a real-world setting, where feature predictions are used for a particular purpose. Due to the extensiveness of the empirical evaluation showcased in this research, we limited the scope of analysis to the feature extraction task, leading to future work on additional quality characteristics from ISO 25010, such as functional appropriateness or interaction capability of T-FREX as a component in a real mobile app review mining pipeline. In addition, the selection of $f_\beta$ is conditioned by the human evaluation results. Different experiments might lead to different $\beta$ values. To reduce the threats imposed by this, we reported extended results including precision, recall and harmonized $f_1$ in addition to $f_\beta$ to facilitate replication in other contexts. Furthermore, we conducted a human evaluation with external annotators using a large set of reviews and multiple annotators (5 per task), increasing the reliability of results by using average values. Finally, our human evaluation uses non-developer participants, which may not fully reflect real-world expertise. To address this, we include precision and recall metrics alongside $f_\beta$ in our analysis for a balanced evaluation.

Concerning \textit{internal validity}, the main threat comes from the ground truth dataset used for evaluation.  Collected reviews from Google Play and AlternativeTo might be biased and conditioned by domain-specific particularities. Furthermore, the formalization of what constitutes a \textit{feature} entails some bias, especially given the inconsistencies found in the literature (see Section~\ref{sec:back-feat-ex}). To mitigate this threat, we relied on and leveraged crowdsourced annotations made by real users in an industrial setting. We argue that our findings are then applicable to the concept of feature as being used in real settings, rather than providing or reusing a synthesized data set. In addition, we focused on the evaluation of one instance selection algorithm, as well as specific data partitions. This selection might have introduced some bias in the instance selection process. We limited the scope of analysis based on literature review and procuring resource optimization, validating the decisions and the design evaluated in this research. However, we acknowledge that other instance selection algorithms might produce similar - or even better - results.

Concerning \textit{external validity}, we identify the generalization of research findings as the most compromised threat. Specifically, evaluation of different datasets, such as reviews from other repositories beyond Google Play, features from other sources beyond AlternativeTo, or even for different mobile app categories, might produce different outcomes. We built on previous, validated work to construct datasets for both fine-tuning and extended pre-training, increasing the confidence in the quality and representativeness of the domain of the data used for evaluation. Furthermore, results for model comparative analysis might not translate with different model architectures even with the same evaluation settings. Additionally, the generalization of the value of extended pre-training in the domain of mobile apps to other tasks (e.g., sentiment analysis) remains unexplored.

Finally, concerning \textit{conclusion validity}, decisions and recommendations concerning best and worst models and settings for each scenario are conditioned and restricted to the results presented in this paper. To this end, for the T-FREX baseline, we designed two different learning scenarios (i.e., in-domain vs. out-of-domain) to assess the validity of T-FREX across different contexts of application. Furthermore, we exhaustively reported all metrics to facilitate decisions based on their applicability. While in this paper, for some settings, we limited the scope of these results (i.e., results for RQ\(_4\) are limited to the best configuration in each setting), we provide the source code for all experiments and all tasks depicted in this research to replicate our evaluation. In addition, we provide data artefacts to verify extended results and complement the analytical insights presented in this research.


\section{Related work}\label{sec:related-work}

We structure related work based on two main areas of research. First, we cover automatic methods for NLP-based feature extraction from mobile app reviews. Second, we focus on empirical research depicting approaches to fine-tune LLMs for domain-specific token classification tasks. 

\subsection{Feature extraction}

Dabrowski et al. recently conducted a systematic literature review in the field of mining app reviews for feedback analysis~\citep{Dabrowski2022}. They identified feature extraction as a key task, for which they also conducted replication and comparative studies using multiple relevant approaches in the field~\citep{DABROWSKI2023102181}. The state-of-the-art in the field is mainly represented by syntactic-based approaches. While the SAFE approach is considered one of the standards~\citep{Johann2017}, there are several related contributions published either as early work~\citep{Iacob2014,Guzman2014,Gu2015}, replication studies~\citep{Shah2019} or even as continual evolutions of the same approach~\citep{DRAGONI20191103}. These methods are based on the use of syntactic properties, such as Part-of-Speech (PoS) tags and syntactic dependencies between elements in a given sentence. Using a pattern-matching approach, syntactic methods look for a series of predefined patterns compliant with typical formulations for a feature. Recent work in the field is still applying these methods for complex opinion mining NLP-based pipelines~\citep{Sutino2019,Song2020450,Kasri2020,Hawari2021,Kumari2022}. 

Recently, there have been some initial proposals based on leveraging LLMs to support the feature extraction task. TransFeatEx~\citep{Motger2023} uses a RoBERTa model for extracting syntactic annotations, to which then a syntactic pattern-matching approach can be applied. KEFE~\citep{Wu2021922} uses PoS pattern-extracted features as input to a BERT model for classifying sentences are potential feature mentions. Their focus is on applying this technique to app descriptions, transferring potential feature matches to user reviews. 

Despite extensive work in the field, several challenges posed by these strategies still remain. Performance on the correctness of extracted features is limited, especially when processing user reviews, leading to substantial noise (i.e., increased false positives) and missed features (i.e., increased false negatives) due to the various writing styles and grammatical inaccuracies found in reviews~\citep{Guzman2014,Johann2017,DRAGONI20191103}. Furthermore, evaluation strategies and ground truth are limited to instructed internal coders~\citep{Johann2017,DABROWSKI2023102181}. Our approach delves into these challenges by leveraging crowdsourced annotations by real users and transferring them into real user reviews. In addition to previous considerations, source code for these solutions is scarce~\citep{Dabrowski2022}, and black-box integration is highly limited due to compatibility restrictions (both for syntactic-based and deep-learning-based solutions). We publish T-FREX models on a collection of HuggingFace models ready to be used either for download or with the HuggingFace Inference API\footnote{https://huggingface.co/docs/api-inference/index}, facilitating reusability and integration with other software components.

\subsection{Token classification with LLMs}

Hou et al. recently conducted a systematic literature review in the field of software engineering practices leveraging LLMs~\citep{Hou2024}. Similarly, context-agnostic literature reviews on the generic use of LLMs index several contributions for token classification and NER tasks~\citep{Naveed2024,Zhao2023,Minaee2024}. Several works showed that Transformer-based models trained for NER on common entities (e.g., locations, dates, names) have showcased promising results with respect to traditional ML methods methods~\citep{Jingjing2019,Souza2020}. On the other hand, domain-specific studies have increasingly focused on leveraging LLMs for token classification, particularly in the medical domain. For instance, Tial et al. evaluated different Transformer-based NER models on free-text eligibility criteria from clinical trials \citep{tian2021transformer}. Liu et al. proposed Med-BERT, a medical-dictionary-enhanced BERT specifically tailored for performing NER on medical records \citep{liu2021med}.

On a closer domain, Malik et al. tested three Transformer models (i.e. BERT, RoBERTa and ALBERT) for software-specific entity extraction \citep{malik2022software}. Tabassum et al. instead fine-tuned BERT for a NER task on 20 fine-grained types of computing programming entities from Stack Overflow posts~\citep{Tabassum2020}. Chen et al. proposed a BERT language representation model for extracting cybersecurity-related terms such as software, organizations and vulnerabilities from unstructured texts~\citep{Chen2021236}. Beyond these studies, token classification models leveraging LLMs in the field of software engineering are scarce. To the best of our knowledge, there is no related work using encoder-only LLMs for token classification in mobile app review mining to extract app-related entities.





\section{Conclusions and future work}\label{sec:conclusions}

In this study, we presented T-FREX, a feature extraction method in the context of mobile app reviews leveraging encoder-only LLMs. We redefined feature extraction as a NER task using crowdsourced annotations generated by real users in a real setting. Empirical evaluation of T-FREX baseline (RQ\(_1\)) showcases the potential of T-FREX with respect to previous approaches, improving both precision and recall of extracted features (H\(_1\)). In addition, extending the pre-training of such models with a large dataset of reviews (RQ\(_2\)) resulted in improved correctness of the predictions in almost all settings (H\(_2\)), with only a few epochs to achieve best results in the majority of scenarios. Furthermore, applying our proposal for feature-oriented instance selection (RQ\(_3\)) not only significantly improves the performance efficiency of the fine-tuning process (H\(_3\)), but also increases the correctness of feature predictions in multiple scenarios, especially in the context of base models. Finally, while the combined use of instance selection and extended pre-training (RQ\(_4\)) is not always the best approach, it still improves the correctness of feature extraction while also improving the performance efficiency in most T-FREX settings. 

As future work, we plan on evaluating the generalization of T-FREX models to other mobile app domains. Specifically, we want to explore its ability to generalize and extract features in emerging, disruptive domains, such as AI-based applications. Furthermore, we plan to explore generalization beyond the scope of mobile applications, such as desktop applications or APIs, and other user-generated documents, such as issues and bug reports. Finally, the functional appropriateness and suitability of T-FREX in a real-world setting remain to be explored, particularly its adoption within review mining pipelines for user feedback analysis. 
In conclusion, we envisage that both the methodological contributions as well as the set of T-FREX LLM instances (fine-tuned and with extended pre-training), which are publicly available, might assist future researchers and practitioners by integrating these models into their own review-based NLP pipelines for opinion mining and decision-making tasks where features are considered a core descriptor. 



\section*{Data Availability Statements}

The source code and full app review datasets required for the replication of all experiments and the full evaluation artefacts are publicly available in the latest release of our GitHub repository: \url{https://github.com/gessi-chatbots/t-frex}. The repository's README file includes references to the models published on HuggingFace, including fine-tuned models for feature extraction and LLMs with extended pre-training in the field of mobile app reviews.

\section*{Conflict of Interest}

The authors declared that they have no conflict of interest.

\section*{Funding}

With the support from the Secretariat for Universities and Research of the Ministry of Business and Knowledge of the Government of Catalonia and the European Social Fund.
This paper has been funded by the Spanish Ministerio de Ciencia e Innovación under project/funding scheme PID2020-117191RB-I00 / AEI/10.13039/501100011033.
This paper has been also supported by FAIR - Future AI Research (PE00000013) project under the NRRP MUR program funded by the NextGenerationEU.

\section*{Ethical Approval}

This study adhered to established ethical principles for research involving human participants. As data collection was fully anonymous and conducted through online platforms, formal ethical approval was not required.

\section*{Informed Consent}

All participants provided informed consent before participating in the study. They were informed about the research purpose, data usage, and their right to withdraw at any time. No personally identifiable information was collected.

\section*{Author Contributions}

\textbf{Quim Motger}: Conceptualization, Methodology, Software, Validation, Formal analysis, Investigation, Data Curation, Writing - Original Draft, Visualization. 
\textbf{Alessio Miaschi}: Conceptualization, Methodology, Validation, Investigation, Resources, Writing - Original Draft. 
\textbf{Felice Dell'Orletta}: Conceptualization, Methodology, Writing - Review \& Editing, Supervision, Funding Acquisition. 
\textbf{Xavier Franch}: Conceptualization, Writing - Review \& Editing, Supervision, Funding Acquisition. 
\textbf{Jordi Marco}: Conceptualization, Writing - Review \& Editing, Supervision. 





\bibliographystyle{spbasic}      
\bibliography{bibliography}   

\end{document}